%% file: main.tex
\documentclass[11pt, a4paper, logo, copyright]{googledeepmind}
\usepackage[dvipsnames]{xcolor} 
\usepackage[authoryear, sort&compress, round]{natbib}
\usepackage[utf8]{inputenc} 
\usepackage[T1]{fontenc}    
\usepackage{hyperref}       
\usepackage{url}            
\usepackage{booktabs}       
\usepackage{amsfonts}       
\usepackage{nicefrac}       
\usepackage{microtype}      
\usepackage{xcolor}         

\usepackage{pifont}
\usepackage{float}
\usepackage{makecell}
\usepackage{algpseudocode}
\usepackage{algorithm}

\usepackage{graphicx}
\usepackage{subcaption}
\usepackage{yfonts}
\usepackage{mathtools}
\mathtoolsset{showonlyrefs=true}
\usepackage{wrapfig}
\usepackage{silence}

\usepackage{enumitem}

\newcommand{\R}{\mathbb{R}}

\newcommand{\E}{\mathbb{E}}

\newcommand{\bwdx}[1]{\tilde x_{#1}}

\newcommand{\pnoise}{p_{\text{noise}}}
\newcommand{\pcond}{p_{\text{cond}}}

\newtheorem*{theorem*}{Theorem}

\newtheorem*{prop*}{Property}

\title{Diffusion Fine-tuning with Rewarded Moment Matching Distillation}

\correspondingauthor{}

\author[*,1]{Alexis Jacq}
\author[*,1]{Guillaume Couairon}
\author[1]{Valentin De Bortoli}
\author[1]{Quentin Berthet}
\author[1]{Arnaud Doucet}
\author[1]{Romuald Elie}

\affil[*]{Equal contributions}
\affil[1]{Google DeepMind}

\begin{abstract}
Distillation and Reinforcement Learning (RL) fine-tuning are the primary pillars of diffusion post-training. While traditionally studied in isolation, the interaction between these phases remains poorly understood, and in particular how fine-tuning impacts the generative quality of distilled models. We introduce Rewarded Moment Matching Distillation (RMMD), a novel framework that simultaneously distills diffusion models and maximizes a reward function.
RMMD preserves the high-fidelity ``naturalness'' characteristic of advanced distillation (such as 8-step Moment Matching) by adapting the sampling loop for on-policy training and repurposing the distillation loss as a proxy for integral KL regularization. By evaluating the FID-Reward Pareto fronts on ImageNet, we demonstrate that RMMD achieves superior trade-offs compared to single-step baselines (DI++) and multi-step competitors (DRaFT, HyperNoise). Finally, we apply RMMD to GenCast, a state-of-the-art weather forecasting model, to distill it while optimizing the Continuous Ranked Probability Score (CRPS) metric. The resulting distilled model achieves a $7.5\times$ speedup while outperforming the teacher model on $93$\% of target weather variables, and being better calibrated. This proves that RMMD scales to complex, high-dimensional scientific domains.

\end{abstract}

\begin{document}

\maketitle

\input{content/rmmd}

\input{content/gencast}

\input{content/conclusion}
\newpage
\bibliography{biblio,gencast_bib}

\appendix

\input{content/appendix_rmmd}

\input{content/appendix_gencast}

\end{document}

%% file: content/rmmd.tex
  \section{Introduction}
  \label{sec:intro}
  
  Diffusion models~\citep{ho2020denoising,song2020score} have become the backbone of
  high-fidelity image synthesis~\citep{esser2024scaling,hoogeboom2025simplerdiffusionsid215}, largely because of their remarkable empirical performance and the fact that their training
  objective reduces to a stable regression problem over corrupted data.
  At inference time, however, generating a single sample requires iterating a
  learned denoiser dozens to hundreds of times, severely limiting throughput.
  Distillation compresses this into one or a few steps~\citep{song2023consistency,boffi2024flow,salimans2024multistep,gengmeanflow2025}, while reward
  fine-tuning steers the model toward downstream objectives such as human
  preferences or safety criteria~\citep{black2023training, fan2023dpok,
  xu2023imagereward,uehara2024understanding,clark2023directly}.
  Both are desirable: a practical model must be fast \emph{and} aligned.
  
  Combining the two is non-trivial.
  Merging distillation and reward fine-tuning into a single training
  phase~\citep{li2024reward, ren2024hyper, luo2024diff} is appealing but fragile:
  reward maximization shifts generated samples out of the teacher's input
  distribution, progressively invalidating the distillation signal.
  Fine-tuning an already distilled model avoids this drift, but optimizing a
  reward over a multi-step chain is memory-intensive, and truncating
  backpropagation introduces bias.
  DRaFT-1~\citep{clark2023directly} is a method that differentiates only through the final step. ReFL~\citep{xu2023imagereward} differentiates through a random step, but
  evaluates the reward on a still-noisy latent, giving an inaccurate gradient.
  HyperNoise~\citep{eyring2025noise} sidesteps chain backpropagation entirely by
  perturbing the initial noise, but is structurally confined to low-frequency
  image changes.
  
  We propose \textbf{Rewarded Moment Matching Distillation (RMMD)}, a two-phase procedure that
  connects distillation and fine-tuning in a principled way.
  In the \emph{first phase}, we distill the base model via Moment-Matching
  Distillation (MMD)~\citep{salimans2024multistep}, which matches intermediate
  denoising distributions along the sampling trajectory, yielding a multi-step
  student that closely tracks the teacher's marginals.
  This model is then \emph{frozen} as a stable distributional reference.
  In the \emph{second phase}, we fine-tune the student with a single-step
  gradient on corrupted \emph{on-policy} samples: we noise a current student
  sample to an intermediate timestep, take one denoising step, and evaluate the
  reward, capturing the full frequency range without multi-step unrolling.
  The central contribution of RMMD is recycling the moment-matching loss as a
  \emph{regularizer} during fine-tuning.
  Rather than a heuristic penalty, this directly penalizes the discrepancy
  between the student's intermediate denoising distributions and those of the
  frozen reference, giving a principled, interpretable knob (the regularization
  weight) to trade off reward against distributional fidelity.
  

Finally, we apply Rewarded Moment Matching to distill a diffusion-based weather model, GenCast \citep{price2023gencast}. GenCast is sampled with 59 backbone evaluations (NFEs) for a single 12h forecast, a high cost which motivates a distillation approach. We apply RMMD to GenCast by considering the CRPS scoring rule as a reward function, which is a distance between the marginals of the generative distribution and the ground truth distribution. We show that optimizing CRPS at a $12$ hours lead time provides consistent improvements of the model up to 7-day forecasts, and a much better dispersion of forecasts compared to MMD distillation alone.

Our main contributions are:

\begin{enumerate}[leftmargin=20pt]
    \item The RMMD procedure, a novel method for diffusion distillation while maximizing a reward function;
    \item empirical results demonstrating that RMMD consistently outperforms DRaFT, HyperNoise, and DI++ on FID--Reward Pareto fronts across diverse reward functions; and
    \item using RMMD to improve GenCast, a state-of-the-art diffusion-based weather forecasting model. With RMMD, the distilled model is $7.5$ times faster than the teacher model while being more accurate on 93\% of the forecast weather variables.
\end{enumerate}

\section{Background and notations}
\subsection{Forward process and inference}
A forward diffusion process progressively corrupts clean data $x_0\in \R^d\sim p_0$ from a target data distribution $p_0$, obtaining purely Gaussian noise $x_1\sim \mathcal{N}(0, I_d)$ according to a schedule such that, conditional upon $x_0$, $x_t = \alpha_t x_0 + \sigma_t \varepsilon \sim \pnoise(x_t|x_0)$ for $t\in(0,1]$ with $\varepsilon\sim \mathcal{N}(0, I_d)$ and $\alpha_t$ and $\sigma_t$ defining the signal-to-noise ratio $\text{SNR}_t =\alpha_t^2/\sigma_t^2$, where $\alpha_0=1, \sigma_0=0$ and $\alpha_1=0, \sigma_1=1$. We denote by $p_t$ the marginal distribution of $x_t$.

In diffusion modeling, a neural network $\Psi$ is trained to predict the denoised data $\E[x_0|x_t]$, given a timestep $t\in (0,1]$ and input $x_t$. This is equivalent to learning the score $s(t,x)=\nabla_x \log p_t(x)=(\alpha_t \E[x_0|x_t=x]-x)/\sigma^2_t$. In this work, a network trained in this fashion plays the role of a {\em teacher} network.
Diffusion modeling leverages this denoising network via the generative process, which consists in sampling $x_1 \sim \mathcal{N}(0, I_d)$, and iteratively for $K$ steps $t=K\delta,...,\delta$ for $\delta=1/K$, denoising $\hat x_0^{(t)} = \Psi(x_t, t)$ and sampling $x_{t - \delta}\sim \pcond(x_{t - \delta}|x_{t}, \hat x_0^{(t)}) = \mathcal{N}(x_{t - \delta}|\mu_t(x_{t}, \hat x_0^{(t)}), \Sigma_t)$
where $\Sigma_t$ and $\mu_t(x_t, \hat x_0^{(t)})$ are functions of $\alpha_t$, $\alpha_{t-\delta}$, $\sigma_t$ and $\sigma_{t-\delta}$~\citep{song2021denoising}. 
Considering the model $\Psi$ as a {\em teacher model}, this mechanism induces a distribution $p_{\text{teacher}}$ for synthetic data, that we can use as a target in distillation. 

\subsection{Distributional distillation}
  Backward sampling in diffusion models requires $K\gg 1$ steps, motivating distillation. A student model $\Phi_\theta$, defined as a deterministic function of $x_t$, a
  timestep $t$, and auxiliary noise $\xi\sim q(\xi)$ (e.g. for dropout), is trained so that its
  output $\hat x_0 = \Phi_\theta(x_t,t,\xi)$ matches the teacher posterior sampled with the original many-step diffusion schedule obtained with $\Psi$:
  \begin{equation}\label{eq:distillation_goal}
      \hat{x}_0|x_t \sim p_\text{teacher}(\cdot|x_t).
  \end{equation}
  The quality of the distilled model depends on how well $\hat{x}_0|x_t$ matches the target distribution $p_\text{teacher}(\cdot|x_t)$. If there is an exact equality of distributions for $t=1$, then the student can perfectly match the teacher in one step. Otherwise, multi-step sampling for a number of steps $K_{student} \ll K$ is often beneficial, as in Moment Matching Distillation~\citep{salimans2024multistep} presented below.
  We write $\frac{\partial \hat x_0}{\partial\theta}$ as shorthand for $\frac{\partial \Phi_\theta(x_t,t,\xi)}{\partial\theta}$, and use $\text{sg}[\cdot]$ for the stop-gradient operator.
  
  \paragraph{Diff-Instruct and DI++.}
  Diff-Instruct~\citep{luo2023diff} fixes $t=1$ and trains a single-step student
  by minimizing an integral KL divergence.
  Given $x_1\sim\mathcal{N}(0,I_d)$, the student predicts $\hat x_0\sim
  p_\theta(\hat x_0|x_1)$; a re-noised sample $x'_s\sim\pnoise(x'_s|\text{sg}[\hat x_0])$
  is then drawn at a uniformly sampled $s\in[0,1]$. Marginally, we have $x'_s \sim p_\theta(x'_s)$. The loss and its gradient are defined as
  \begin{align}
      \mathcal{L}_\text{DI}(\theta)
          &= \E_{s}\!\big[w(s)\,\mathrm{KL}
             \!\big(p_\theta(x'_s)\|p_\text{teacher}(x'_s)\big)\big],\\
      \nabla_\theta\mathcal{L}_\text{DI}(\theta)
          &= \E_{s, x'_s}\!\bigg[
             w(s)\!\left(\frac{\partial\hat x_0}{\partial\theta}\right)^{\!\top}
             \!\big(s_\text{student}(s, x'_s)-s_\text{teacher}(s, x'_s)\big)\bigg],
  \end{align}
  where $s_\text{student}$ and $s_\text{teacher}$ are the score functions of the
  student and teacher respectively.
  DI++~\citep{luo2024diff} augments this with a differentiable reward term:
  \begin{equation}
      \nabla_\theta\mathcal{L}_\text{DI++}(\theta)=
      \E_{s,x_1,\hat x_0,x'_s}\!\bigg[
      w(s)\!\left(\frac{\partial\hat x_0}{\partial\theta}\right)^{\!\top}
      \!\big(s_\text{student}(s, x'_s)-s_\text{teacher}(s, x'_s)
      -\lambda\nabla_{\hat x_0}R(\hat x_0)\big)\bigg],
  \end{equation}
  where $\lambda$ balances reward and KL regularization.
  The student may optionally be pre-distilled with Diff-Instruct, after which the
  KL term acts as a classical regularizer keeping the generator close to its
  initial distribution.
  
  \paragraph{Moment-Matching Distillation (MMD).}
  MMD extends distillation to arbitrary timesteps
  $t\sim\mathrm{Uniform}[0,1]$ by matching generalized moments of student and
  teacher posteriors~\citep{salimans2024multistep}:
  \begin{equation}\label{eq:lossSUR}
      \mathcal{L}_\text{MMD}(\theta)=
      \E_{t,s,x_t,\hat x_0,x'_s}\!\big[
      \hat x_0^\top\!\textup{sg}\big(m_\text{student}(s, x'_s)-m_\text{teacher}(s, x'_s)\big)\big],
  \end{equation}
where the expectation is w.r.t. $t\sim\mathrm{Uniform}[0,1]$, $x_t \sim p_t$, a student prediction $\hat x_0\sim p_\theta(\hat
  x_0|x_t)$, and a re-noised sample $x'_s\sim p_{\textup{cond}}(x'_s|x_t,\text{sg}[\hat x_0])$ at
  $s\in[t{-}\delta_\text{student},t)$. Here $m_\text{teacher}(s,x'_s)=\E_{p(x_0|x'_s)}[x_0|x'_s]$ is predicted by the
  teacher and $m_\text{student}(s, x'_s)=\E_{p_\theta(\hat x_0|x'_s)}[\hat x_0|x'_s]$
  by an auxiliary network trained alongside the student to predict $\hat x_0$ from $x'_s$. We detail the auxiliary's loss in Appendix\ref{app:auxiliary}. Since $x'_s$ uses a stop-gradient on $\theta$, the loss gradient is
  \begin{equation}\label{eq:mmd_grad}
      \nabla_\theta\mathcal{L}_\text{MMD}(\theta)=
      \E_{t,s,x_t,\hat x_0,x'_s}\!\bigg[
      \!\left(\frac{\partial\hat x_0}{\partial\theta}\right)^{\!\top}
      \!\big(m_\text{student}(s, x'_s)-m_\text{teacher}(s, x'_s)\big)\bigg].
  \end{equation}
  When $m_\text{student}=m_\text{teacher}$ everywhere, student and teacher
  marginals coincide~\citep{salimans2024multistep}.
  Training specializes to step size $\delta_\text{student}$, requiring a number of sampling steps of
  $K_\text{student}=1/\delta_\text{student}\ll K$ for the student, using DDPM as for the teacher.
 We refer the reader to ~\cite{salimans2024multistep} for a complete presentation of MMD.

  \section{Rewarded Moment Matching}
  \label{sec:rmm}

\begin{figure}[ht]
  \begin{center}
  \includegraphics[width=0.95\linewidth]{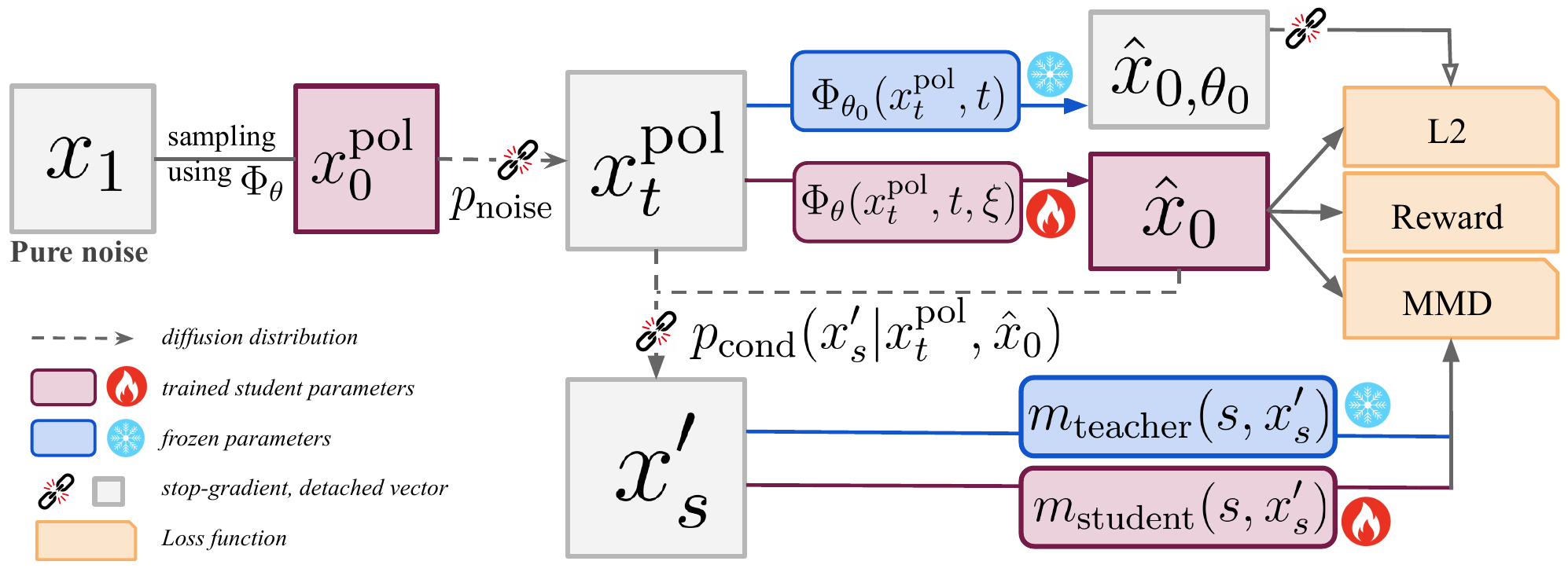}
  \caption{RMMD: an on-policy student sample $x^\text{pol}_0$ is re-noised to
  $x^\text{pol}_t$, from which the student predicts $\hat x_0$.
  The loss combines a reward term $R(\hat x_0)$ with the moment-matching
  regularizer $\mathcal{L}_\text{MMD}$ and the L2 regularizer $\mathcal{L}_\text{L2}$.}
  \label{fig:rmm}
  \end{center}
  \vspace{-1em}
  \end{figure}

\subsection{From off-policy to on-policy reward optimization.}
  A natural baseline is to add a reward term directly to the MMD gradient
  (eq.~\eqref{eq:mmd_grad}):
  \begin{equation}
      \nabla_\theta\mathcal{L}_\text{naive}(\theta)=\E_{t, s, x_t,\hat x_0 ,x'_s}\bigg[
      \!\left(\frac{\partial\hat x_0}{\partial\theta}\right)^{\!\top}
      \!\big(m_\text{student}(s, x'_s)-m_\text{teacher}(s, x'_s)
      - \lambda\nabla_{\hat x_0} R(\hat x_0)\big)\bigg].
  \end{equation}
  This is analogous to DI++, but introduces two problem:
  First, $\hat x_0$ is predicted from a re-noised \emph{off-policy} data point
  $x'_s$, so an extra dataset is needed and fine-tuning is capped by the best
  reward achievable on the data.
  Second, as the student distribution shifts toward high-reward outputs, the
  data-marginal $p_t$ becomes an increasingly poor approximation of the
  student's own intermediate distribution.
  
We address both issues by sampling $x_t$ not from the noisy data $p_t$ but from noisy on-policy samples $x^\text{pol}_t \sim p_{\text{noise}}(x_t | \text{sg}[\bwdx{0}])$ where $\bwdx{0}\sim p_\theta$ is obtained by sampling the student with $K_{\text{student}}$ steps.

With this marginal, the reward objective now directly approximates the student's expected reward:
  \begin{equation}
  \label{eq:reward_objective}
  \mathcal{L}_\text{Reward}(\theta)
  =\E_{t,\,x^\text{pol}_t,\,\xi}
  \big[R\big(\Phi_\theta(x^\text{pol}_t,t,\xi)\big)\big]
  =\E_{\hat x_0}
  \big[R\big(\hat x_0\big)\big]
  \simeq \E_{\bwdx{0}\sim p_\theta}\big[R(\bwdx{0})\big]
  \end{equation}
  The quality of the approximation depends on the quality of the distilled model in Equation~\eqref{eq:distillation_goal}. With the stop-gradient on $\bwdx{0}$, the gradient is given by 
  \begin{equation}
  \label{eq:reward_grad}
  \nabla_\theta\mathcal{L}_\text{Reward}(\theta) =
  \E_{\hat x_0}
  \bigg[\!\left(\frac{\partial\hat x_0}{\partial\theta}\right)^{\!\top}
  \nabla_{\hat x_0}R(\hat x_0)\bigg].
  \end{equation}

\subsection{RMMD objective}

We combine the on-policy reward gradient (Equation \eqref{eq:reward_grad}) with the on-policy version of MMD regularization (Equation~\eqref{eq:mmd_grad}), yielding the on-policy gradient
  \begin{equation}
  \label{eq:rmm}
  \nabla_\theta\mathcal{L}_\text{RMMD}(\theta)=
  \E_{\substack{t, s,\,x^\text{pol}_t,
  \,\hat{x}_0, x'_s}}\!\bigg[
  \!\left(\frac{\partial\hat x_0}{\partial\theta}\right)^{\!\top}
  \!\big(m_\text{student}(s, x'_s)-m_\text{teacher}(s, x'_s)
  -\lambda\nabla_{\hat x_0}R(\hat x_0)\big)\bigg].
  \end{equation}
  The reference moment $m_\text{teacher}$ can be replaced by the frozen auxiliary
  model learned during the initial distillation phase. The process is illustrated in Figure \ref{fig:rmm}.
  In contrast to DI++, this regularization is not equivalent to an integral KL
  divergence unless $x'_s$ is drawn from the forward diffusion (rather than the
  conditional), in which case $\mathcal{L}_\text{MMD}$ recovers the
  score-matching objective of DI as shown in \cite{salimans2024multistep}.

On top of the MMD distillation objective and reward objective, we also add a regularization loss that penalizes the student if its predictions are significantly different from the MMD-distilled model $\Phi_{\theta_0}$, with the loss
\begin{equation}
\label{eq:l2}
    \mathcal{L}_\text{L2}(\theta)=
    \mathbb{E}_{t,x^\text{pol}_t,\xi}\!\big[||\Phi_\theta(x^\text{pol}_t,t,\xi)-\Phi_{\theta_0}(x^\text{pol}_t,t)||^2\!\big],
\end{equation}
where $\Phi_{\theta_0}$ is used without noise $\xi$.

\label{sec:multi-rmmd}
Finally, we extend the RMMD objective ~\eqref{eq:rmm} to reward functions that depend on multiple samples, e.g. for measuring diversity. We can sample $(\hat{x}_0^{(i)})_i$ and combine the associated MMD regularization losses with $R(\hat{x}_0^{(1)}, \hat{x}_0^{(2)}, ..)$. 

  \section{Related Work}
  \label{sec:related}
  
  \textbf{Diffusion model distillation.}
  A large body of work has focused on reducing the number of sampling steps
  required by diffusion models. Consistency Models~\citep{song2023consistency, song2024improved} and extensions like Consistency Trajectory Models~\citep{kim2024consistency} and Shortcut Models~\citep{frans2025one} enforce self-consistency along
  the probability-flow ODE, enabling one- or few-step generation. MeanFlow~\citep{geng2025mean} learns an average velocity field for one-step generation, and the improved iMF variant addresses training stability and recovers inference flexibility \citep{geng2025imf}.
  Multistep Consistency Models~\citep{heek2024multistep} extend Consistency Models by
  matching intermediate denoising marginals, yielding high-quality multi-step
  students that closely track the teacher. Score-distillation methods such as Diff-Instruct~\citep{luo2023diff} minimize
  an Integral KL divergence between teacher and student score functions. In this work, we use Moment Matching Distillation~\citep{salimans2024multistep}, which is a multi-step distillation method based on stochastic sampling that provides samples with image quality similar to the teacher when using 8 steps.

   \textbf{Reward fine-tuning of diffusion models.}
  Several methods directly optimize a differentiable reward along the sampling
  trajectory. DI++~\citep{luo2024diff} builds on top of the Diff-Instruct framework and also supports
  reward fine-tuning from a pre-distilled initialization.
  DRaFT~\citep{clark2023directly} backpropagates through $K$ denoising steps,
  while DPOK~\citep{fan2023dpok} and DDPO~\citep{black2023training} cast the
  chain as a Markov decision process and apply policy-gradient methods.
  ReFL~\citep{xu2023imagereward} backpropagates through a single randomly chosen
  step, which can be suboptimal as the reward is evaluated on a noisy intermediate
  prediction.
  Implicit Diffusion~\citep{marion2024implicit} frames fine-tuning as stochastic
  optimal control, enabling gradient computation through stochastic samplers.
  All of these methods operate on undistilled models and therefore incur the full
  multi-step inference cost during training.
  
\textbf{Joint distillation and reward fine-tuning.}
  RG-LCD~\citep{li2024reward}, Hyper-SD~\citep{ren2024hyper}, and
  DI++~\citep{luo2024diff} augment distillation training with a reward term,
  using the distillation objective as an implicit regularizer.
  A shared limitation is that online fine-tuning causes generated samples to
  drift outside the teacher's distribution, gradually degrading the distillation
  signal.
  Reward-Instruct~\citep{luo2025rewardinstruct} sidesteps distillation altogether
  by directly fine-tuning a few-step teacher to maximize a reward, a strategy
  sensitive to the expressiveness of the reward function.
  
  \textbf{Fine-tuning distilled models.}
  The closest line of work to ours fine-tunes an already distilled model in a
  separate phase.
  HyperNoise~\citep{eyring2025noise} trains a lightweight network to perturb noise
  inputs so that they follow a reward-shifted distribution, avoiding the memory
  cost of multi-step backpropagation.
  Because the perturbation acts only at the noise level, HyperNoise is constrained
  to low-frequency modifications, which can be insufficient for reward functions
  affecting high frequency structure.
  Our work shares the two-phase spirit of HyperNoise but differs in two key
  respects: we backpropagate through single-step predictions on \emph{corrupted
  on-policy samples} rather than perturbing noise inputs, and we explicitly
  regularize fine-tuning with the moment-matching loss, providing a principled
  connection between the distillation and fine-tuning phases.  
  
\section{Experiments}
In this section, we first evaluate RMMD on ImageNet~\citep{deng2009imagenet} with the U-Vit backbone from Simple Diffusion~\citep{hoogeboom2023simple,hoogeboom2025simplerdiffusionsid215} and some selected reward functions. We justify the use of a multi-step regime for fine-tuning by comparing RMMD with DI++ \citep{luo2024diff}, and then compare our method to other existing multi-step such as DRaFT \citep{clark2023directly}\ and HyperNoise \citep{eyring2025noise} on selected reward functions. Finally, we use RMMD to distill and improve the state-of-the-art diffusion-based weather model GenCast \citep{price2023gencast}.

\subsection{Fine-tuning and evaluation details}

\textbf{Distillation.} The first stage of RMMD is MMD distillation \citep{salimans2024multistep} without any reward optimization. We use MMD with 8 sampling steps, which we found to be a very strong baseline, and tuned some hyper-parameters to start from the strongest possible distilled model. For the fine-tuning phase, we keep the same hyper-parameters as for distillation (same batch sizes, optimizers, learning rates) except that we use 10,000 steps and decay the learning rate to zero. We optimize the combined loss $\mathcal{L} = \mathcal{L}_{\textup{MMD}}^\text{online} - \lambda R + \lambda_{\textup{reg}} \mathcal{L}_{\text{L2}}$, for different values of $\lambda$ and fixing $\lambda_{\textup{reg}} = \lambda / 2$.

\textbf{Evaluation.} Since there can be a trade-off between the two objectives of optimizing a reward function and preserving image quality, all our evaluations are based on FID-vs-Reward Pareto fronts. We fine-tune models with $J$ reward factors $\{\lambda_j\}_{j=1}^J$ and evaluate them every 2500 steps during training, evaluating the FID~\citep{heusel2017gans} and Reward on $50,000$ generated samples. The FID-Reward Pareto front is obtained as the set of Pareto-optimal (FID, Reward) evaluation results.

\textbf{Reward functions.} We evaluate RMMD with simple reward functions to validate the method, and then focus on more challenging real-case scenarios like weather forecasting. The black-and-white reward is given by the pixel-wise distance between an image and its black-and-white version, averaging over the color channels. The Laplacian smoothing reward is given by the average distances between a pixel and its four neighbours (up, left, down, right). IS reward directly uses the Inception Score~\citep{barratt2018note} as a reward, and CLIP-red is the CLIP~\citep{radford2021learning} alignment score with the word ``red''.

\subsection{Strong distilled student with 8-steps sampling}
\begin{figure}[ht]
\includegraphics[width=\linewidth]{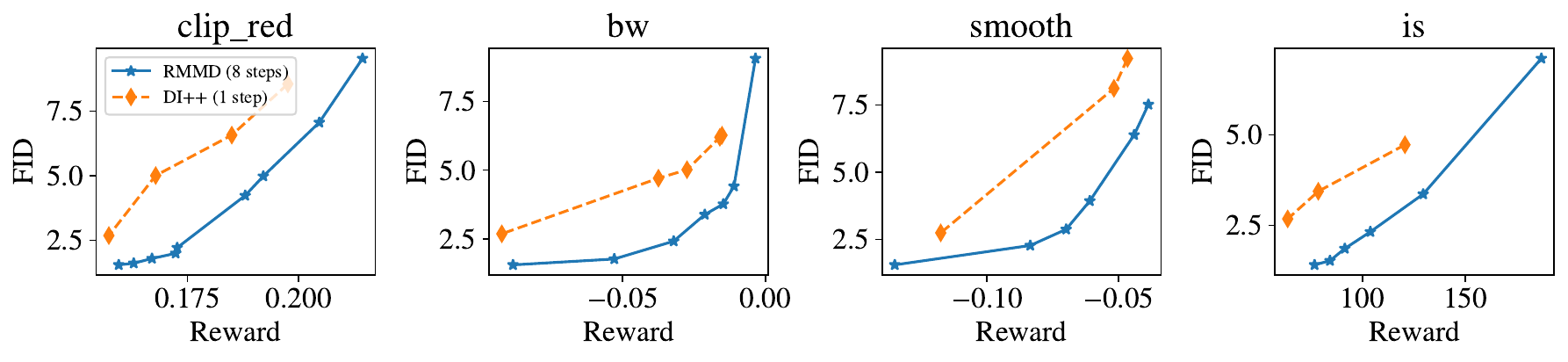}
\caption{The advantage of multi-step sampling: FID-Reward Pareto obtained with RMMD (8 steps) and DI++ (1 step).}
\label{fig:dipp}
\end{figure}

In the first stage, we employ models distilled using 8-step sampling, which provides an optimal balance between inference speed and image quality, evidenced by an FID score of 1.26 on ImageNet $64 \times 64$, closely matching the teacher’s score of 1.19. Furthermore, multi-step distillation significantly benefits RMMD; we validate this in Figure \ref{fig:dipp} by comparing our approach against the 1-step DI++ method. For a fair comparison, we trained the strongest possible 1-step baseline, achieved by distilling a model into 2 steps via MMD before further distilling it into 1 step with DI++ (FID 2.65). As shown in Figure \ref{fig:dipp}, our 8-step RMMD consistently outperforms the 1-step DI++ alternative.
\vspace{-1em}

\subsection{Comparing to other multi-step fine-tuning methods}
\begin{figure}[ht]
\includegraphics[width=\linewidth]{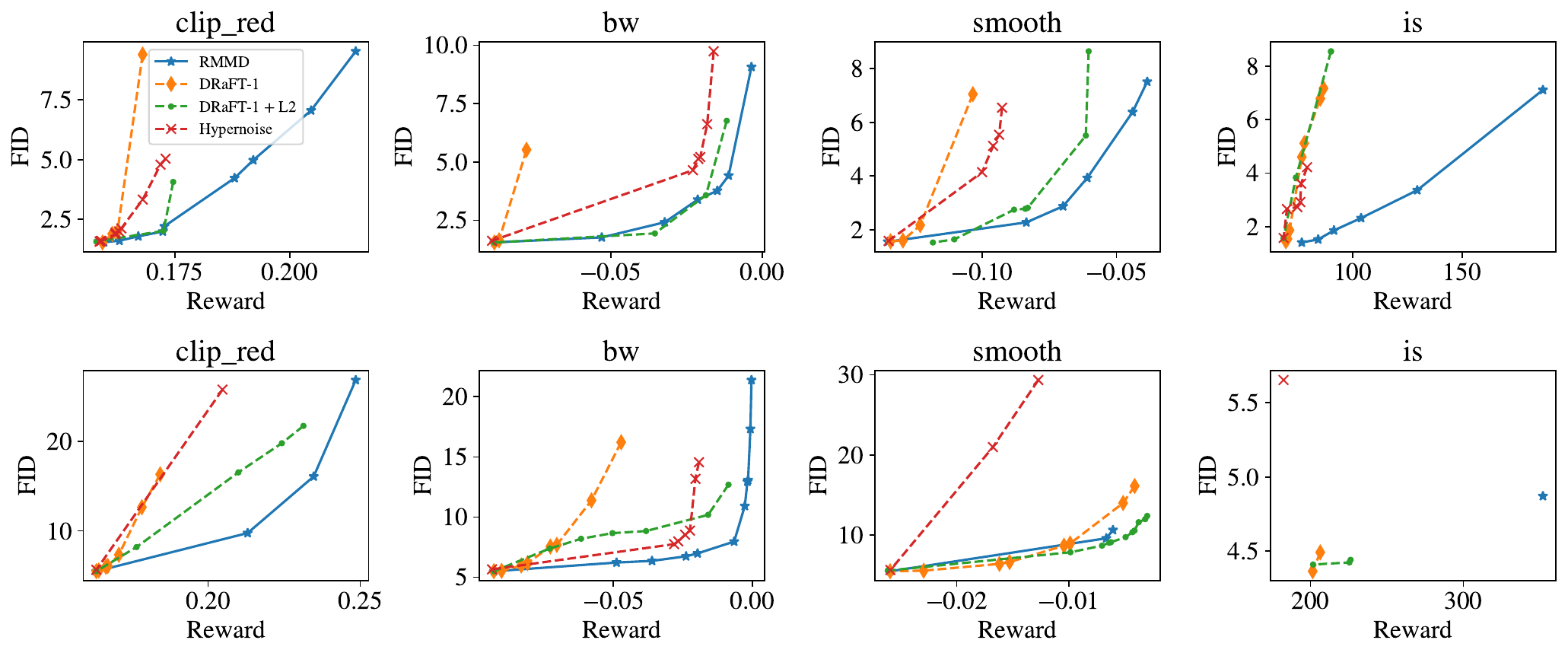}
\caption{FID-Reward Pareto for different multi-step fine-tuning methods. Top: ImageNet-64 images generated in 8 steps. Bottom: ImageNet-512 images generated in 2 steps.}
\label{fig:baselines}
\end{figure}

We compare the reward fine-tuning stage of RMMD with HyperNoise~\citep{eyring2025noise} and DRaFT~\citep{clark2023directly}, at both small (64 x 64) and large (512 x 512) image resolution, and both 2-step and 8-step regimes, in Figure \ref{fig:baselines}. For DRaFT, we found that adding the L2 regularization leads to much better generations than just using LoRA in few-step distilled regimes (this regularization is called ``KL regularization'' in DRaFT~\citep{clark2023directly}, since it can be viewed as optimizing an integral KL divergence in non-distilled networks). In all settings, our fine-tuning method led to better Pareto fronts, especially in neural-network based rewards (CLIP alignment and Inception score).

\vspace{-1em}

\subsection{Qualitative comparison}
\begin{figure}

    \begin{subfigure}{0.19\textwidth}
        \includegraphics[width=\textwidth]{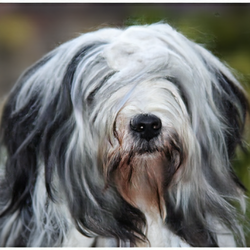}
    \end{subfigure}
    \hfill 
    \begin{subfigure}{0.19\textwidth}
        \includegraphics[width=\textwidth]
        {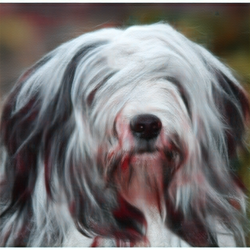}
    \end{subfigure}
    \hfill 
    \begin{subfigure}{0.19\textwidth}
        \includegraphics[width=\textwidth]
        {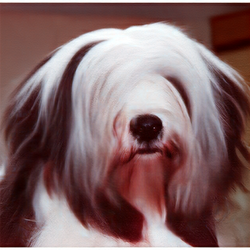}
    \end{subfigure}
    \hfill 
    \begin{subfigure}{0.19\textwidth}
        \includegraphics[width=\textwidth]
        {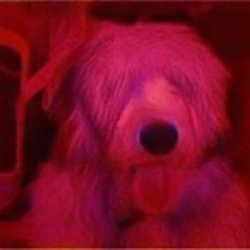}
    \end{subfigure}
    \hfill 
    \begin{subfigure}{0.19\textwidth}
        \includegraphics[width=\textwidth]
        {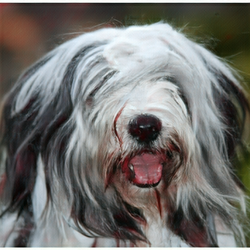}
    \end{subfigure}
    \hfill 
    \vspace{1mm}

    \begin{subfigure}{0.19\textwidth}
        \includegraphics[width=\textwidth]{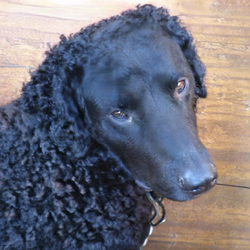}
    \end{subfigure}
    \hfill 
    \begin{subfigure}{0.19\textwidth}
        \includegraphics[width=\textwidth]
        {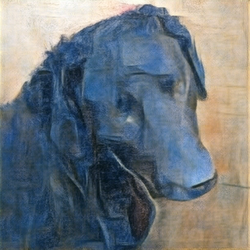}
    \end{subfigure}
    \hfill 
    \begin{subfigure}{0.19\textwidth}
        \includegraphics[width=\textwidth]
        {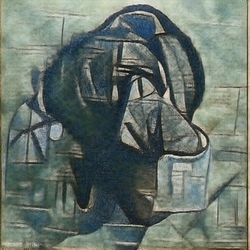}
    \end{subfigure}
    \hfill 
    \begin{subfigure}{0.19\textwidth}
        \includegraphics[width=\textwidth]
        {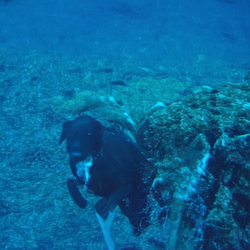}
    \end{subfigure}
    \hfill 
    \begin{subfigure}{0.19\textwidth}
        \includegraphics[width=\textwidth]
        {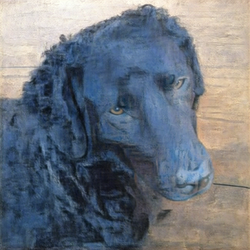}
    \end{subfigure}
    \hfill 
    \vspace{1mm}

    \begin{subfigure}{0.19\textwidth}
        \includegraphics[width=\textwidth]{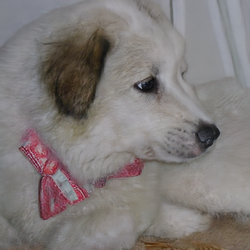}
        \caption{Teacher}
    \end{subfigure}
    \hfill 
    \begin{subfigure}{0.19\textwidth}
        \includegraphics[width=\textwidth]
        {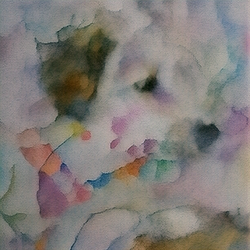}
        \caption{DRaFT-1}
    \end{subfigure}
    \hfill 
    \begin{subfigure}{0.19\textwidth}
        \includegraphics[width=\textwidth]
        {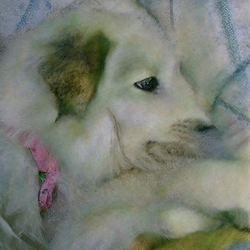}
        \caption{DRaFT-2}
    \end{subfigure}
    \hfill 
    \begin{subfigure}{0.19\textwidth}
        \includegraphics[width=\textwidth]
        {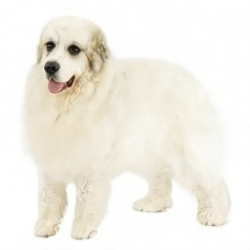}
        \caption{HyperNoise}
    \end{subfigure}
    \hfill 
    \begin{subfigure}{0.19\textwidth}
        \includegraphics[width=\textwidth]
        {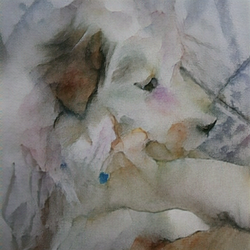}
        \caption{\textbf{RMMD (Ours)}}
    \end{subfigure}
    \hfill 
    \vspace{1mm}

\caption{Visualization of fine-tuning behaviors using a 2-step ImageNet512 MMD teacher. Across all examples, we start from the same initial noise $x_1$ and denoise using identical random seeds and class conditioning. \textbf{The target rewards are CLIP alignments for the concepts 'red', 'Picasso', and 'watercolor'}. While \textbf{DRaFT-1} preserves the original semantic content, it introduces adversarial artifacts that exploit CLIP features (e.g., red shadows, cubic shapes, and patchy textures) and reduces overall sharpness. \textbf{DRaFT-2} causes drastic distribution shifts and significantly deteriorates image quality. \textbf{HyperNoise} struggles to optimize complex style rewards like Picasso and watercolor, instead defaulting to broad color shifts (blue and white, respectively). In contrast, \textbf{RMMD} (Ours) successfully integrates subtle modifications to increase the reward without significantly deviating from the original data distribution. Further examples are provided in the Appendix.}
\label{fig:qualitative}
\end{figure}

Figure~\ref{fig:qualitative} shows samples obtained by fine-tuning over the same pre-distilled 2-step ImageNet512 model using the different multi-step methods.
In each case, we start from the same initial noise $x_1$, and denoise using the same random seed, conditioning on the bullfinch bird class. DRaFT-1 adaptations only affect high frequency changes without modifying the content. DRaFT-2 (back-propagating the gradient on the ``whole" 2-step sampling) is prone to reward hacking (blindly coloring everything in red for the CLIP alignment with 'red'). HyperNoise tends to move further away from the teacher distribution for similar reward. RMMD reaches the best trade-off between image quality and reward.

%% file: content/gencast.tex
\subsection{Application of RMMD to Weather forecasting with GenCast distillation}

In this section, we use RMMD to improve the GenCast weather model, based on diffusion \citep{price2023gencast}. Weather forecasting consists in predicting the evolution of some physical variables (temperature, geopotential, humidity, wind) over time, starting from an initial condition. A gridded historical dataset of these fields, called ERA5, is publicly available \citep{hersbach2020era5} and is used to train machine learning-based weather models such as GenCast. Given the value of weather variables $x^t$ at a given time $t$, GenCast is trained to learn the distribution of these variables 12 hours later $p(x^{t+\delta} | x^t, x^{t-\delta})$. This transition distribution is modeled with a conditional diffusion model, requiring 59 function evaluations (NFE) for sampling. To produce forecasts at a longer horizon, the model is rolled out auto-regressively. More details about GenCast are in Appendix \ref{app:gencast}.

To improve GenCast with RMMD, we use the Continuous Ranked Probability Score (CRPS) scoring rule \citep{ferro2014fair} as a reward function to optimize. CRPS measures the compatibility of a probabilistic forecast with an observation, and is minimum when the forecast follows the exact same distribution as the observation. It is computed separately for each dimension of the data and then averaged. Given an input state $(x^t, x^{t-\delta})$, we use sample $M$ predictions $(\hat{x}^{t+\delta, i})^M_{i=1}$, called \textit{members}, from the generative model $p_\theta(x^{t+\delta} | x^t, x^{t-\delta})$. We also denote by $x^{t+\delta}$ the single sample from the ground truth distribution available in the dataset. The sample-based unbiased estimator for CRPS is then defined, for dimension $r$ and $\tau = t+\delta$, as 
\begin{equation}
    \mathrm{CRPS}_r(\{\hat{x}^{\tau,i}\}_{i=1}^M, x^{\tau}) = \frac{1}{M} \sum_{i=1}^M |\hat{x}^{\tau, i}_{(r)} - x^{\tau}_{(r)}| - \frac{1}{2M(M-1)} \sum_{i=1}^M \sum_{j=1}^M |\hat{x}^{\tau,i}_{(r)} - \hat{x}^{\tau, j}_{(r)}|
\end{equation}

In theory, explicit CRPS optimization would be unnecessary if the generative model was perfectly recovering the true transition distribution $p(x^{t+\delta} | x^t, x^{t-\delta})$. However, in practice, diffusion models often suffer from under-dispersion, failing to capture the full variance of possible outcomes, and one of the strengths of RMMD is allowing us to address this problem.




    




\paragraph{Experimental details.} For our experiments, we use a GenCast teacher that was trained for 500k with a total batch size of 128, on 1 degree resolution maps (resulting in 180$\times$360 latitude-longitude maps) available on GitHub \citep{graphcast_github}.
We then run the MMD algorithm for 300k steps with a total batch size of 16, which corresponds to ~7.5\% of the initial compute budget.
We then continue distillation with RMMD (with CRPS), with a second phase of 300k optimization steps at a batch size of 16. CRPS being a multi-sample reward function, we use the 2-sample variant of RMMD presented in section \ref{sec:multi-rmmd} and use a global reward weighting of $\lambda=0.3$. The final model can be sampled with 8 steps of DDPM sampling instead of 59 for the teacher, resulting in a~$7.5\times$ speed-up. For evaluation, we take each initialization date of the evaluation year (2018), roll out our model for 15 steps (maximum lead time of 7.5 days) with $M=8$ members, and compute CRPS separately at each lead time, following standard procedure.

\paragraph{Evaluation metrics.}
Given a reference ground truth trajectory $x^{t+\delta:t+K\delta}$, we can measure how well the distribution of a set of samples $(\hat{x}^{t+\delta:t+K\delta, i})_{i=1}^M$ matches the true distribution over trajectories with CRPS. The CRPS is computed separately for each physical variable $v$ and lead time $\tau = t + k\delta$.

In the remainder of the paper, we report relative CRPS improvement of a model over the default GenCast teacher model. These CRPS improvements are averaged over physical variables and lead times $\tau = t + k\delta$ for $k \in \{1, \dots, K\}$, to give a representative summary of a model's performance called ``Rel. CRPS score'', e.g. in Table \ref{tab:summary}. We also define the ``Win Rate'' as the fraction of physical variables and lead times for which the model is better than the teacher model.

We also use Spread-skill ratio~\citep{fortin2014should} to measure whether the forecasts are correctly calibrated, a value $<1$ indicating an under-dispersed forecast and a value $>1$ indicating an over-dispersed forecast; see Appendix \ref{app:metrics} for a definition.

\begin{wraptable}[12]{r}{0.5\columnwidth}
    \centering
    \small
    \vspace{-2em}
    \begin{tabular}{l|r|r}
        Name & \makecell{CRPS \\ improv.(↑)} & \makecell{Win \\ rate(↑)} \\
        \hline
        Teacher & 0\% & N/A \\
        Plain MMD & -1.32\% & 4.9\% \\
        MMD (best) & 0.82\% & 75.0\% \\
        RMMD w/ CRPS (offline) & 1.11\% & 89.2\% \\
        RMMD w/ CRPS (online) & 1.51\% & 93.0\% \\

    \end{tabular}
    \caption{Summary table of ablations. Our best MMD model is $7.5\times$ faster than the GenCast model, while being better on 75\% of variables. The RMMD with CRPS (online) model is also $7.5\times$ faster and better on 93\% of variables.}
    \label{tab:summary}
\end{wraptable}

\paragraph{Evaluation of RMMD.} We present an overview of our results in Table \ref{tab:summary}, where we compare the following models to the GenCast teacher (sampled with default parameters and 59 NFE): (i) A model distilled with MMD without any change to the original algorithm; (ii) A second model distilled with MMD by optimizing some hyperparameters (using $\eta=0.5$ in DDIM sampling and changing the diffusion noise schedule to $\rho=100)$. It has an average CRPS improvement of 0.8\% and is better than the teacher on 75\% of variables.

Two models are distilled with RMMD and CRPS, (i) a first version where $x^t_0$ is sampled offline from the dataset; (ii) A second version where $x^\text{pol}_0$ is sampled from the current policy. This is more costly (about $2\times$) but is also more effective (+0.4\% CRPS improvements, +3.8\% Win Rate.). There are two main advantages of the on-policy variant: (i) the network is trained on noisy versions of its own generations, which is closer to the distribution of its inputs during sampling; (ii) For a given $(x^{t-\delta},x^t)$, there is only one transition $x^t \rightarrow x^{t+\delta}$ in the training set; on-policy allows us to sample more transitions with the current policy, which can reduce overfitting.

\begin{figure}[htbp]
    \centering
    \includegraphics[width=\linewidth]{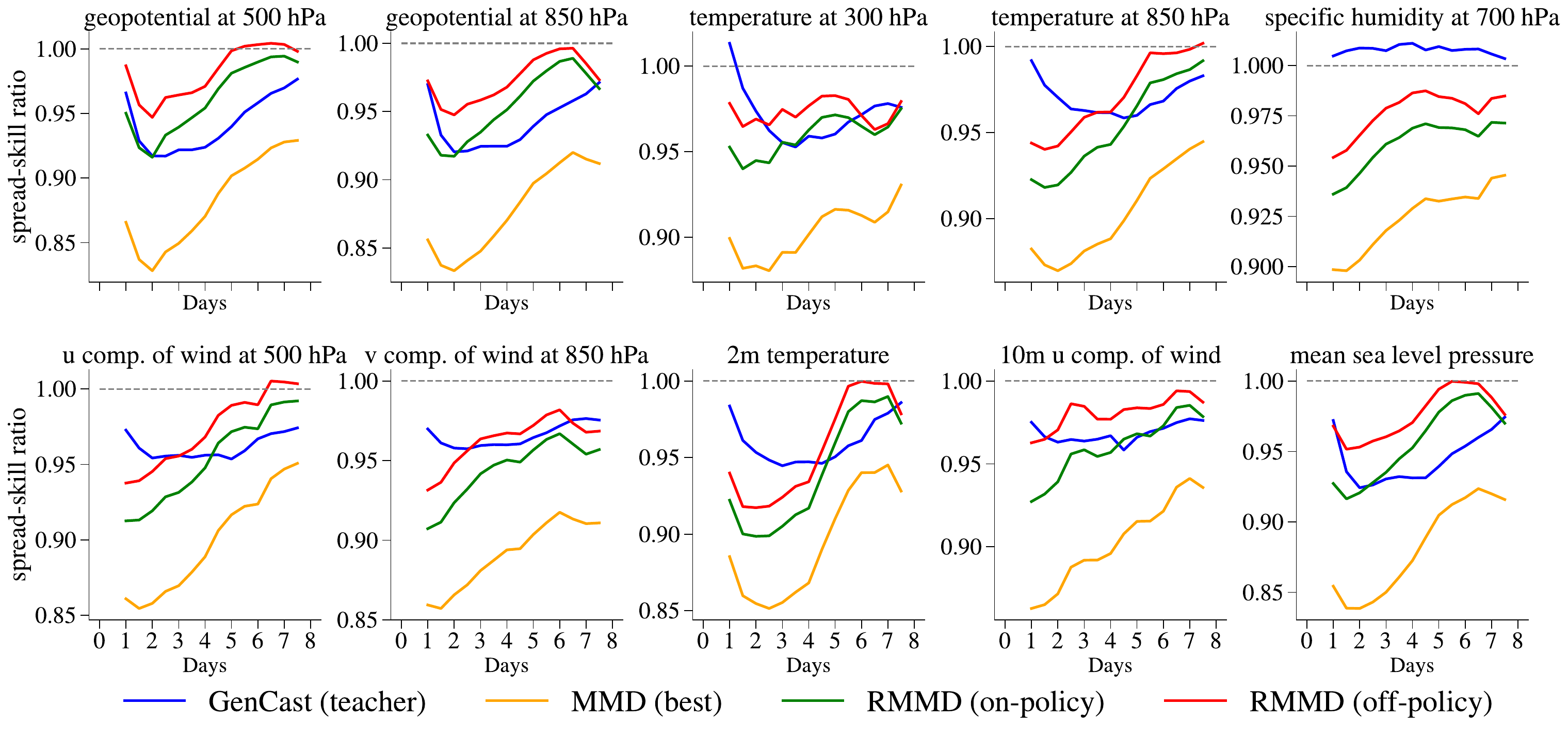}
    \caption{Spread-skill ratio of our RMMD-finetuned models. Optimizing CRPS greatly improves dispersion of generated weather states compared to using MMD alone, generally matching or improving GenCast's dispersion, except for humidity.}
    \label{fig:spskr_mmr}
\end{figure}

\paragraph{RMMD improves model calibration.}
In Figure \ref{fig:spskr_mmr}, we observe that CRPS optimization also improves dispersion (measured by spread-skill ratio) compared to the reference MMD model used for initialization. However surprisingly, the on-policy version, which has better CRPS scores, has slightly worse dispersion compared to the off-policy model.

\paragraph{Discussion on the reward objective.} CRPS is the metric on which weather forecasts are evaluated at different lead times $(t+k\delta)_{k\in[1..K]}$ along trajectories, and it is also the metric that is being optimized with RMMD. Therefore it could be seen as obvious that optimizing CRPS will improve evaluation metrics. However, we only optimize the next-state CRPS at a $12$ hours time difference and we observe that the relative gain of the distilled model over the teacher actually increases for larger lead times in auto-regressive rollouts (see Figure \ref{fig:crps_mmron} in Appendix), and the CRPS at larger lead times is not being optimized. We believe this shows that RMMD improves modeling of the distribution, because reward hacking for the next-state predictions would not yield improvements in auto-regressive rollouts.

%% file: content/conclusion.tex
\subsection{Limitations}
The quality of generated samples is inherently limited by the performance of the distilled model prior to reward optimization. While RMMD successfully maintains the generation quality of the initial MMD-distilled model, it does not improve FID during the reward optimization phase. This limitation motivated our choice of a highly robust starting point: an $8$-step MMD distillation.

Because RMMD is an on-policy method (consistent with DRaFT-1 or HyperNoise), it necessitates sampling from the current policy at every training step. By utilizing $8$ sampling steps with a stop-gradient on on-policy samples, we observed only a $2\times$ computational slowdown compared to an off-policy variant. This trade-off is justified by a significant gain in accuracy, as measured on GenCast distillation. Finally, RMMD requires a differentiable reward function; it cannot optimize non-differentiable or black-box rewards without the use of a differentiable surrogate.

\section{Conclusion}

We develop a fine-tuning method that leverages a model initially distilled via MMD, and optimizes differentiable rewards while using an on-policy version of the moment-matching loss for regularization. Our empirical evaluations on ImageNet demonstrate that RMMD shows superior trade-offs compared to single-step methods like DI++, ensuring higher generation quality through retained multi-step capabilities. Furthermore, it leads to better Pareto fronts than other multi-step fine-tuning approaches such as DRaFT and HyperNoise across the considered reward functions, ranging from simple pixel-level metrics to complex semantic signals like CLIP alignment. 

We have applied RMMD to weather forecasting with the distillation of the diffusion-based model GenCast, and demonstrated a $7.5\times$ speed improvement while improving the model's probabilistic predictions on $93$\% of variables. RMMD also solved the under-dispersion issue of MMD distillation. This paves the way for better diffusion-based models in science where there is a differentiable objective that can be optimized.

%% file: content/appendix_rmmd.tex
\section{Additional details}

For all experiments on ImageNet, we use the U-Vit backbone from Simple Diffusion~\citep{hoogeboom2023simple,hoogeboom2025simplerdiffusionsid215}. Our only modifications are to allow for a dropout rate of $0.1$ in all transformer blocks.
We use a pixel space based diffusion process, and, for 64 x 64 images, we use a shifted cosine schedule for distillation with a logSNR shift $b=\log(2)$~\citep{hoogeboom2025simplerdiffusionsid215}, which we found to bring notable improvement even with a teacher trained with a symmetric cosine schedule, as we can see in  Table~\ref{tab:dropout}. With this schedule, the SNR for timestep $t=0.5$ is equal to 2 instead of 1 for the default cosine schedule.

\subsection{Training the auxiliary model to predict the student's moment}\label{app:auxiliary}

We follow~\citep{salimans2024multistep} for training the auxiliary model alongside the student to predict the student's moment $m_{\text{student}}(s, x'_s)$. For an auxiliary model $m_{\text{student}} = \Phi_{\theta_\text{aux}}$ (initialized with $\theta_0$), the loss is written:

\begin{equation}
    \mathcal{L}_\text{auxiliary}(\theta_\text{aux}) = \mathbb{E}_{t, \hat{x}_0, x'_s}\!\big[||\Phi_{\theta_\text{aux}}(x'_s) - \hat{x}_0||^2 + ||\Phi_{\theta_\text{aux}}(x'_s) - \Phi_{\theta_0}(x'_s)||^2\!\big],
\end{equation}
where the first term is a regression to predict the student's generation $\hat{x}_0$ (at noise level $t$), given the next denoising step $x'_s$. The second term is a L2 regularization to ensure the auxiliary weights stay close to the initial ones.

\subsection{Hyperparameters}

Here is the list of hyper-parameters that we use for RMMD, on the ImageNet64, ImageNet512 and ERA5 datasets:
\begin{table}[ht]
\begin{tabular}{l|ccc}
                                      & ImageNet 64x64                                                                                                                              & ImageNet 512x512                                                                                                                            & ERA5 1º                                                                                                                                     \\ \hline
Architecture                          & U-ViT                                                                                                                                       & U-ViT                                                                                                                                       & Graph Transformer                                                                                                                           \\
Teacher Model                         & \begin{tabular}[c]{@{}c@{}} Simpler Diffusion \\ (SiD2)\end{tabular}                                                                                                 & \begin{tabular}[c]{@{}c@{}} Simpler Diffusion \\ (SiD2)\end{tabular} & GenCast                                                                                                                                     \\
Dropout (MMD)                              & 0.1                                                                                                                                         & 0.1                                                                                                                                         & 0.0                                                                                                                                         \\
Dropout (RMMD)                              & 0.1                                                                                                                                         & 0.0                                                                                                                                         & 0.0                                                                                                                                         \\
MMD steps                             & 50k                                                                                                                                         & 50k                                                                                                                                         & 300k                                                                                                                                        \\
RMMD steps (phase 2)                  & 10k                                                                                                                                         & 10k                                                                                                                                         & 300k                                                                                                                                        \\
Batch size                            & 2048                                                                                                                                        & 2048                                                                                                                                        & 16                                                                                                                                          \\
Training hardware & 16 TPU-v5 & 16 TPU-v5 & 16 TPU-v6 \\
fine-tuning samples         & 120M                                                                                                                                        & 120M                                                                                                                                        & 9.6M                                                                                                                                        \\
Data augmentation                     & Random hflip                                                                                                                      & Random hflip                                                                                                                     & None                                                                                                                                        \\
Optimizer                             & \begin{tabular}[l]{l@{}l@{}l@{}}Adam($\beta_1=0.9$, \\ $\beta_2=0.99$, \\ $\epsilon=1e-12$)\end{tabular} & \begin{tabular}[l]{l@{}l@{}l@{}}Adam($\beta_1=0.9$, \\ $\beta_2=0.99$, \\ $\epsilon=1e-12$)\end{tabular} & \begin{tabular}[l]{l@{}l@{}l@{}}Adam($\beta_1=0.9$, \\ $\beta_2=0.99$, \\ $\epsilon=1e-12$)\end{tabular} \\
gradient accumulation                 & 1                                                                                                                                           & 1                                                                                                                                           & 8                                                                                                                                           \\
Learning Rate                         & 1e-5                                                                                                                                        & 1e-5                                                                                                                                        & 1e-7                                                                                                                                        \\
Reward weight $\lambda$               & variable                                                                                                                                    & variable                                                                                                                                    & 0.3                                                                                                                                         \\
Reg. weight $\lambda_{reg}$ & $\lambda/2$                                                                                                                                 & $\lambda/2$                                                                                                                                 & 1                                                                                                                                           \\
Noise schedule                        & Cosine                                                                                                                                      & Cosine                                                                                                                                      & EDM w/ $\rho=100$ \\
DDPM epsilon                          & 1                                                                                                                                           & 1                                                                                                                                           & 0.5                                                                                                                                         \\
\end{tabular}
\end{table}

\subsection{Comparison against competing methods}

We evaluate FID-Reward Pareto fronts by sweeping over reward scaling factor $\lambda$. For DRaFT  \citep{clark2023directly} with LoRA, we sweep over LoRA weighting coefficients instead of reward factors; for DRaFT with LoRA mixed with L2 regularization $\lambda_{reg}$, we fix the best LoRA coefficient and sweep over $\lambda_{reg}$).

\section{Additional experiments}

\subsection{Alternative sampling}

\begin{figure}[ht]
\includegraphics[width=\linewidth]{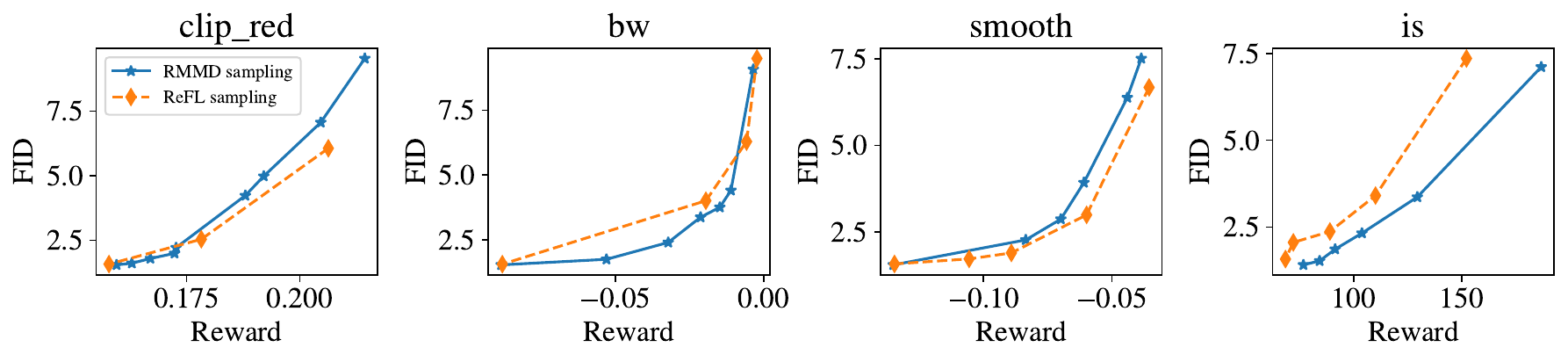}
\caption{FID-Reward Pareto obtained with methods for sampling $x^\text{pol}_t$: ReFL (discrete) sampling and continuous sampling. The lower and righter the better.}
\label{fig:refl}
\end{figure}

Another strategy to sample pathwise $x^\text{pol}_t$, inspired by ReFL~\citep{xu2023imagereward} consists in early stopping the denoising process at a random step $t_{stop}\in \{0,\delta_\text{student}, \dots, (1-K_\text{student})\delta_\text{student} \}$. The advantage is that it only trains at the time steps that matter, \textit{i.e.} the $\check{x}_t$ seen at inference are the same as during training. This however limits the generalization of the moments matched by MMD, which performs better on continuous time steps $t$~\citep{salimans2024multistep}. We implemented and compared both sampling methods in Figure~\ref{fig:refl}, and observed slightly better results with early stopping. Unfortunately, since our implementation operates on batches, we had to either use the same stopping time for a whole batch, or do the whole sampling and keep all time steps in memory to randomly draw a batch of different times (the latter solution led to the reported results, but is much more memory expensive).

Figure~\ref{fig:refl} compares the Pareto fronts obtained with the two sampling methods (in both cases we apply both RMMD and L2 regularization). The discrete sampling performs slightly better on the CLIP-based reward and the smoothness, while the continuous sampling works slightly better with the Inception Score reward.

\subsection{Evaluating L2 regularization}

In this section, we evaluate a baseline that uses reward maximization along with L2 regularization, but without the MMD distillation objective.

One risk with L2 regularization alone is that it can reduce diversity, since
\begin{equation}
\mathbb{E}[||\hat x_{0,\theta}-\hat x_{0,\theta_0}||^2|\bwdx{t}]=\mathrm{Tr}(\mathrm{cov}[\hat x_{0,\theta}|\bwdx{t}])+\mathrm{Tr}(\mathrm{cov}[\hat x_{0,\theta_0}|\bwdx{t}])+||\mathbb{E}[\hat x_{0,\theta}|\bwdx{t}]-\mathbb{E}[\hat x_{0,\theta_0}|\bwdx{t}]||^2,
\end{equation}
where $\hat x_{0,\theta}=\Phi_{\theta}(\bwdx{t},t,\xi)$ is learned and $\hat x_{0,\theta_0}=\Phi_{\theta_0}(\bwdx{t},t)$ is the target. In our implementation, only the mean of $p_\theta(\hat x_0|\bwdx{t})$ is modeled and the variability is obtained by a dropout in the weights, so we avoid this side effect. In a setting where the variance is learned, a regularization based on a scoring rule~\citep{de2025distributional} instead could lead to better results.
The discrete ReFL sampling combined with only this L2 regularization would be equivalent to the ReFL method introduced in~\citep{xu2023imagereward}, fine-tuning over a MMD-distilled model. 

\begin{figure}[ht]
\includegraphics[width=\linewidth]{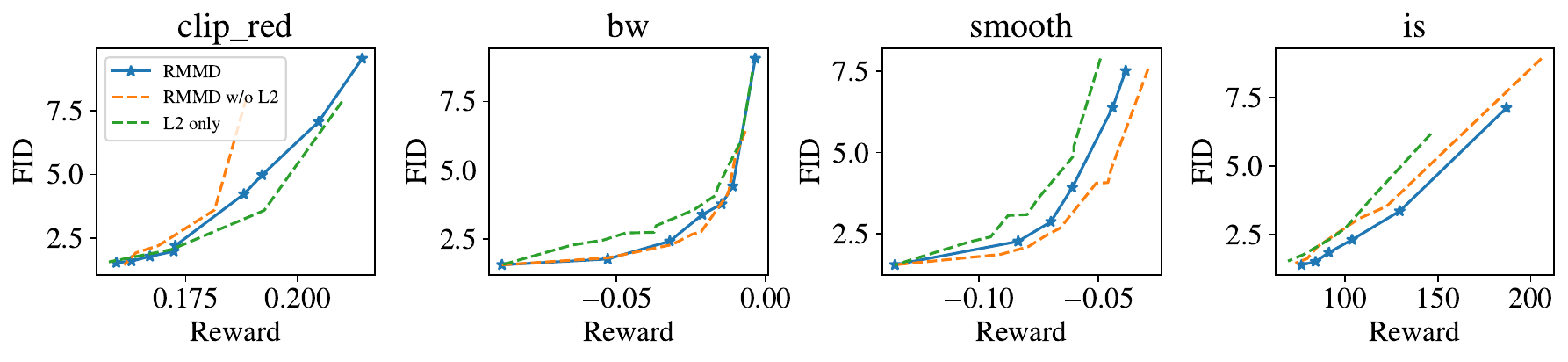}
\caption{FID-Reward Pareto for different regularization techniques: Rewarded Moment Matching without L2 regularization, L2 regularization (L2) and RMMD (mix of both moment matching and L2 regularization). The lower and righter the better. Using both forms of regularization improve the FID-Reward Pareto fronts.}
\label{fig:reg}
\end{figure}

Figure~\ref{fig:reg} shows FID-Reward Pareto obtained with the two regularization approaches on 4 different reward functions. These experiments are conducted at image resolution 64 x 64 and at the 8-step regime. In pixel-wise rewards (black-and-white and smoothness), MMD regularization is better than L2 regularization, but the opposite is observed in the CLIP alignment reward. The combination MMD regularization and L2 regularization stays closer to the best Pareto for these three rewards, and outperforms both isolated methods for the Inception score. 

\subsection{Reward functions}

\begin{figure}
    \begin{subfigure}{0.19\textwidth}
        \includegraphics[width=\textwidth]{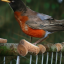}
    \end{subfigure}
    \hfill 
    \begin{subfigure}{0.19\textwidth}
        \includegraphics[width=\textwidth]{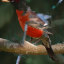}
    \end{subfigure}
    \hfill 
    \begin{subfigure}{0.19\textwidth}
        \includegraphics[width=\textwidth]{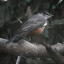}
    \end{subfigure}
    \hfill 
    \begin{subfigure}{0.19\textwidth}
        \includegraphics[width=\textwidth]{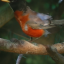}
    \end{subfigure}
    \hfill 
    \begin{subfigure}{0.19\textwidth}
        \includegraphics[width=\textwidth]{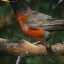}
    \end{subfigure}
    \hfill 
    \begin{subfigure}{0.19\textwidth}
        \includegraphics[width=\textwidth]{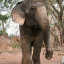}
        \caption{Base}
    \end{subfigure}
    \hfill 
    \begin{subfigure}{0.19\textwidth}
        \includegraphics[width=\textwidth]{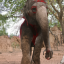}
        \caption{CLIP 'red'}
    \end{subfigure}
    \hfill 
    \begin{subfigure}{0.19\textwidth}
        \includegraphics[width=\textwidth]{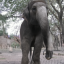}
        \caption{Black and white}
    \end{subfigure}
    \hfill 
    \begin{subfigure}{0.19\textwidth}
        \includegraphics[width=\textwidth]{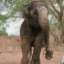}
        \caption{Laplacian smooth}
    \end{subfigure}
    \hfill 
    \begin{subfigure}{0.19\textwidth}
        \includegraphics[width=\textwidth]{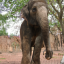}
        \caption{Inception Score}
    \end{subfigure}
\caption{Changes, for a similar seed and class, when maximizing the different reward functions, while staying at a decent FID level (FID <= 8) on ImageNet64.}
\label{fig:rewards}
\end{figure}

Figure~\ref{fig:rewards} displays the resulting changes when maximizing the different reward functions, while staying at an FID below 8 on ImageNet 64. For a given class, the denoising processes start from the same initial noise and sample each step using the same random seed.

\subsection{Effect of dropout on MMD}

We report in Table~\ref{tab:dropout} the FID of generations obtained by adding stochasticity in the student's network prediction via a dropout on our implementation of MMD (only during first distillation phase), as well as shifting the schedule for the 64x64 resolution.

\begin{table}[ht]
    \centering
    \begin{tabular}{c|c|c|c}
        & I64 - 8 steps &  I64 - 2 steps  &  I512 - 2 steps \\
    \hline
        MMD (paper) & 1.24 & 3.86 & -\\
    \hline
        MMD (ours) & 1.35 & 2.0 & 9.7\\
    \hline
        MMD + dropout & 1.37 & 1.66 & 5.4 \\
    \hline
        MMD + dropout + shift & 1.26 & 1.4 & - \\
    \end{tabular}
    \caption{Effect of dropout and shifted schedule on our implementation of MMD on the FID.}
    \label{tab:dropout}
\end{table}

\subsection{Empirical comparison of DRaFT, HyperNoise and RMMD}

\begin{figure}
    \begin{subfigure}{0.19\textwidth}
        \includegraphics[width=\textwidth]{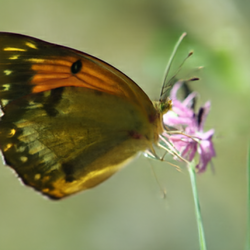}
    \end{subfigure}
    \hfill 
    \begin{subfigure}{0.19\textwidth}
        \includegraphics[width=\textwidth]{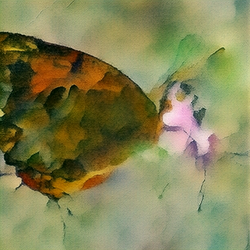}
    \end{subfigure}
    \hfill 
    \begin{subfigure}{0.19\textwidth}
        \includegraphics[width=\textwidth]{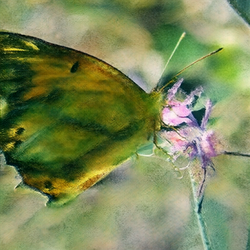}
    \end{subfigure}
    \hfill 
    \begin{subfigure}{0.19\textwidth}
        \includegraphics[width=\textwidth]{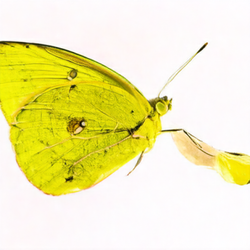}
    \end{subfigure}
    \hfill 
    \begin{subfigure}{0.19\textwidth}
        \includegraphics[width=\textwidth]{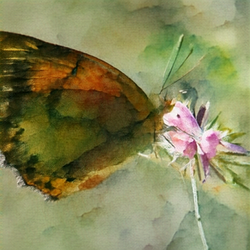}
    \end{subfigure}
    \hfill 
    \vspace{1mm} 

    \begin{subfigure}{0.19\textwidth}
        \includegraphics[width=\textwidth]{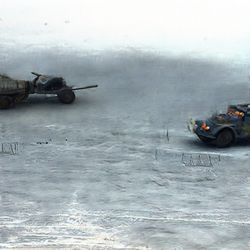}
    \end{subfigure}
    \hfill 
    \begin{subfigure}{0.19\textwidth}
        \includegraphics[width=\textwidth]{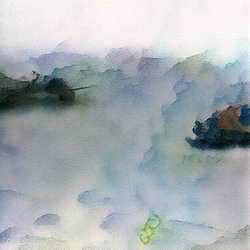}
    \end{subfigure}
    \hfill 
    \begin{subfigure}{0.19\textwidth}
        \includegraphics[width=\textwidth]{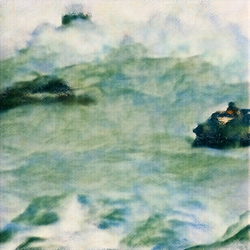}
    \end{subfigure}
    \hfill 
    \begin{subfigure}{0.19\textwidth}
        \includegraphics[width=\textwidth]{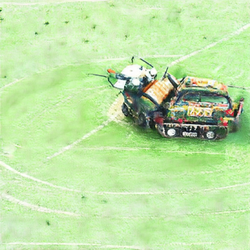}
    \end{subfigure}
    \hfill 
    \begin{subfigure}{0.19\textwidth}
        \includegraphics[width=\textwidth]{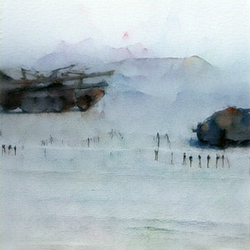}
    \end{subfigure}
    \hfill 
    \vspace{1mm} 

    \begin{subfigure}{0.19\textwidth}
        \includegraphics[width=\textwidth]{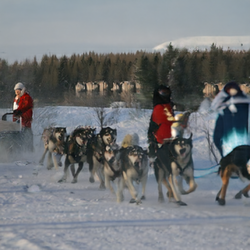}
    \end{subfigure}
    \hfill 
    \begin{subfigure}{0.19\textwidth}
        \includegraphics[width=\textwidth]{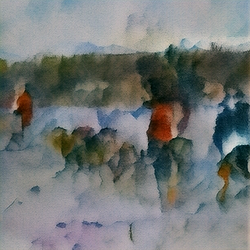}
    \end{subfigure}
    \hfill 
    \begin{subfigure}{0.19\textwidth}
        \includegraphics[width=\textwidth]{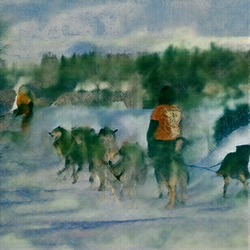}
    \end{subfigure}
    \hfill 
    \begin{subfigure}{0.19\textwidth}
        \includegraphics[width=\textwidth]{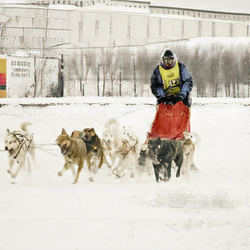}
    \end{subfigure}
    \hfill 
    \begin{subfigure}{0.19\textwidth}
        \includegraphics[width=\textwidth]{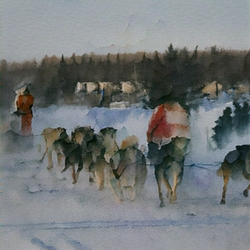}
    \end{subfigure}
    \hfill 
    \vspace{1mm} 

    \begin{subfigure}{0.19\textwidth}
        \includegraphics[width=\textwidth]{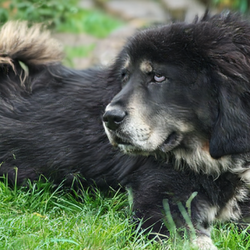}
        \caption*{teacher}
    \end{subfigure}
    \hfill 
    \begin{subfigure}{0.19\textwidth}
        \includegraphics[width=\textwidth]{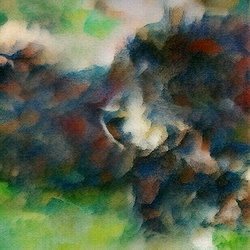}
        \caption*{DRaFT-1}
    \end{subfigure}
    \hfill 
    \begin{subfigure}{0.19\textwidth}
        \includegraphics[width=\textwidth]{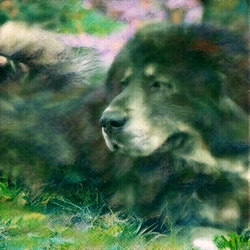}
        \caption*{DRaFT-2}
    \end{subfigure}
    \hfill 
    \begin{subfigure}{0.19\textwidth}
        \includegraphics[width=\textwidth]{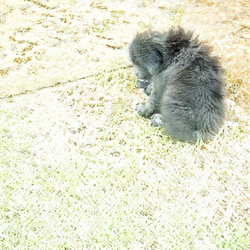}
        \caption*{HyperNoise}
    \end{subfigure}
    \hfill 
    \begin{subfigure}{0.19\textwidth}
        \includegraphics[width=\textwidth]{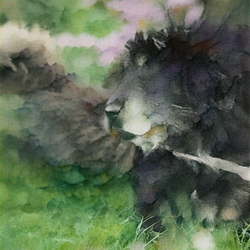}
        \caption*{\textbf{RMMD (ours)}}
    \end{subfigure}
    \hfill 
    \vspace{1mm} 

\caption{More samples of i512 image generated in 2 diffusion steps, using the teacher model, RMMD, DRaFT-1 or DRaFT-2, optimizing for the CLIP alignment with 'Watercolor'}
\label{fig:watercolor}
\end{figure}

\begin{figure}
    \begin{subfigure}{0.19\textwidth}
        \includegraphics[width=\textwidth]{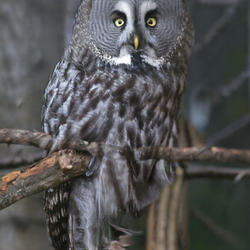}
    \end{subfigure}
    \hfill 
    \begin{subfigure}{0.19\textwidth}
        \includegraphics[width=\textwidth]{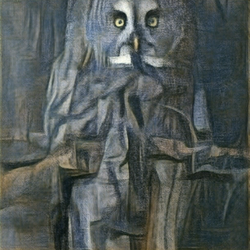}
    \end{subfigure}
    \hfill 
    \begin{subfigure}{0.19\textwidth}
        \includegraphics[width=\textwidth]{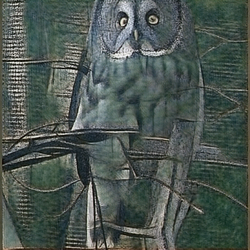}
    \end{subfigure}
    \hfill 
    \begin{subfigure}{0.19\textwidth}
        \includegraphics[width=\textwidth]{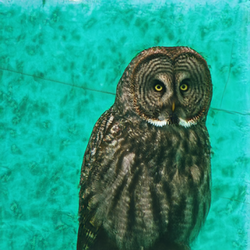}
    \end{subfigure}
    \hfill 
    \begin{subfigure}{0.19\textwidth}
        \includegraphics[width=\textwidth]{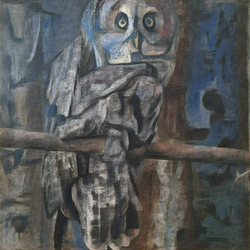}
    \end{subfigure}
    \hfill 
    \vspace{1mm} 

    \begin{subfigure}{0.19\textwidth}
        \includegraphics[width=\textwidth]{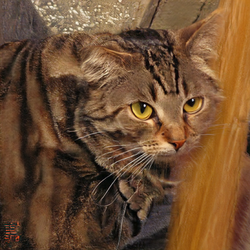}
    \end{subfigure}
    \hfill 
    \begin{subfigure}{0.19\textwidth}
        \includegraphics[width=\textwidth]{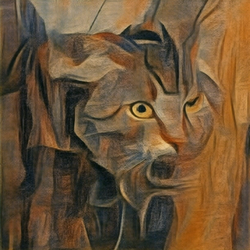}
    \end{subfigure}
    \hfill 
    \begin{subfigure}{0.19\textwidth}
        \includegraphics[width=\textwidth]{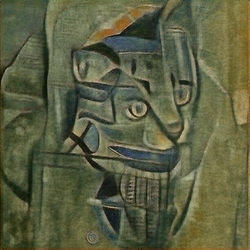}
    \end{subfigure}
    \hfill 
    \begin{subfigure}{0.19\textwidth}
        \includegraphics[width=\textwidth]{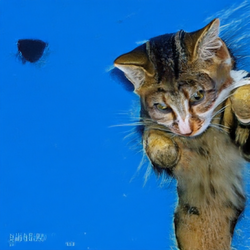}
    \end{subfigure}
    \hfill 
    \begin{subfigure}{0.19\textwidth}
        \includegraphics[width=\textwidth]{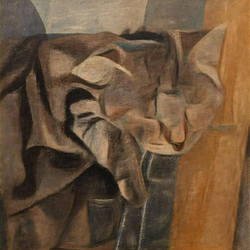}
    \end{subfigure}
    \hfill 
    \vspace{1mm} 

    \begin{subfigure}{0.19\textwidth}
        \includegraphics[width=\textwidth]{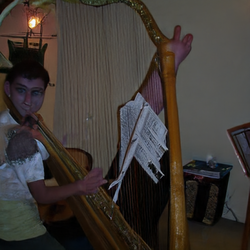}
    \end{subfigure}
    \hfill 
    \begin{subfigure}{0.19\textwidth}
        \includegraphics[width=\textwidth]{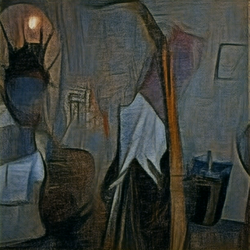}
    \end{subfigure}
    \hfill 
    \begin{subfigure}{0.19\textwidth}
        \includegraphics[width=\textwidth]{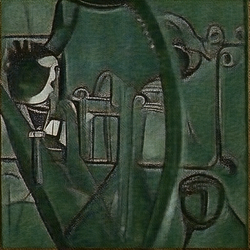}
    \end{subfigure}
    \hfill 
    \begin{subfigure}{0.19\textwidth}
        \includegraphics[width=\textwidth]{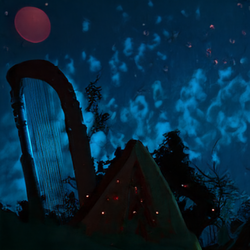}
    \end{subfigure}
    \hfill 
    \begin{subfigure}{0.19\textwidth}
        \includegraphics[width=\textwidth]{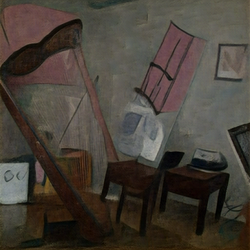}
    \end{subfigure}
    \hfill 
    \vspace{1mm} 

    \begin{subfigure}{0.19\textwidth}
        \includegraphics[width=\textwidth]{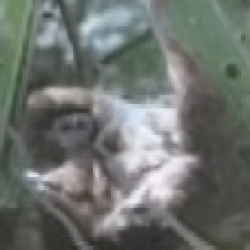}
        \caption*{teacher}
    \end{subfigure}
    \hfill 
    \begin{subfigure}{0.19\textwidth}
        \includegraphics[width=\textwidth]{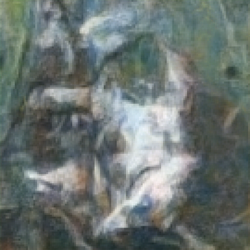}
        \caption*{DRaFT-1}
    \end{subfigure}
    \hfill 
    \begin{subfigure}{0.19\textwidth}
        \includegraphics[width=\textwidth]{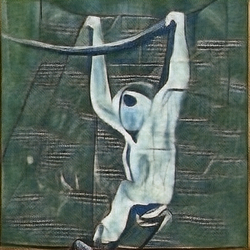}
        \caption*{DRaFT-2}
    \end{subfigure}
    \hfill 
    \begin{subfigure}{0.19\textwidth}
        \includegraphics[width=\textwidth]{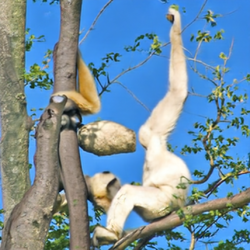}
        \caption*{HyperNoise}
    \end{subfigure}
    \hfill 
    \begin{subfigure}{0.19\textwidth}
        \includegraphics[width=\textwidth]{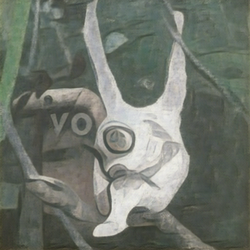}
        \caption*{\textbf{RMMD (ours)}}
    \end{subfigure}
    \hfill 
    \vspace{1mm} 

\caption{More samples of i512 image generated in 2 diffusion steps, using the teacher model, RMMD, DRaFT-1 or DRaFT-2, optimizing for the CLIP alignment with 'Picasso'}
\label{fig:picasso}
\end{figure}

We provide more samples obtained with the different fine-tuning methods on Fig.~\ref{fig:watercolor} (CLIP alignment with the word 'watercolor') and Fig.~\ref{fig:picasso} (CLIP alignment with the word 'Picasso').

%% file: content/appendix_gencast.tex
\section{Additional information on the distillation of GenCast}

\subsection{Definition of metrics} \label{app:metrics}

In this section, we provide the metrics that we use for evaluating our fine-tuned weather models.

\paragraph{CRPS.} Given an input state $x^t$, the generative models predict a distribution $p(x^{t+\delta} | x^t)$ as a set of $M$ samples $(\hat{x}^{t+\delta, i})_i$ and a single sample $x^{t+\delta}$ from the ground truth distribution is available in the dataset. The CRPS~\citep{ferro2014fair} for dimension $r$ and verification time $\tau = t + k\delta$ is then defined as

\begin{equation}
    \text{CRPS}(\{\hat{x}^{\tau,i}\}_i, x^{\tau}) = \frac{1}{M} \sum_{i=1}^M |\hat{x}^{\tau, i} - x^{\tau}| - \frac{1}{2M(M-1)} \sum_{i=1}^M \sum_{j=1}^M |\hat{x}^{\tau,i} - \hat{x}^{\tau, j}|
\end{equation}

The CRPS is then averaged for each initialization time $t$.

\paragraph{Ensemble Mean RMSE.} The Ensemble Mean RMSE is the error between the ground truth and the average over members. It is defined for each physical variable $v$ and lead time $k$ as

\begin{align*}
\text{EnsMeanRMSE}_{k,v} &= \sqrt{\frac{1}{T} \sum_t \Vert y^{t+k\delta}  - \frac{1}{M} \sum_i x^{t+k\delta, i} \Vert^2}
\end{align*}

\paragraph{Spread-Skill Ratio.} The Spread Skill ratio~\citep{fortin2014should} is another metric to assess whether the set of samples has the correct dispersion or spread. The (bias-corrected) spread-skill ratio is defined as 

\begin{align*}
\text{Spread}_k &= \sqrt{\frac{1}{T} \sum_t \frac{1}{M} \sum_i\Vert x^{t+k\delta, i} - \frac{1}{M} \sum_j x^{t+k\delta, j}\Vert^2} \\
\text{SpreadSkillRatio}_k &= \sqrt{\frac{M+1}{M}} \frac{\text{Spread}_k}{\text{EnsMeanRMSE}_k}
\end{align*}

The reasoning is as follows: under the assumption of a perfect ensemble forecast, the ensemble mean should have the same distance on average to the ground truth or to a random ensemble member, hence the spread-skill ratio should be equal to 1. If it is below one, the spread is too small (relative to the model's error) and the model is said to be "underdispersive"; if it is above one, the model is "overdispersive".

\subsection{The GenCast model} \label{app:gencast}

GenCast is a model trained to predict the transition distribution $p(x^{t+\delta}|x^t, x^{t-\delta})$ with $\delta=12h$. $x^t$ contains six variables sampled at different altitude levels in the atmosphere (parameterized by pressure): temperature (T), humidity (Q), Geopotential (Z), and the three components of the wind vector (U, V, W). These variables are sampled at pressure levels ranging from 50 hPa in the upper atmosphere to 1000 hPa close to the Earth's surface. In addition, $x_t$ also contains surface variables, temperature at 2-meters (t2m), mean sea-level pressure (msl), u and v component of wind at 10meters (10u and 10v). We use a version of GenCast trained at a 1º resolution, resulting in equirectangular maps of size $180\times 360$. With all the variables included, A state $x^t$ is of dimension 5,313,600.

\section{Additional experiments of RMMD for GenCast}
\subsection{Score card of RMMD model}

\begin{figure}[ht]
    \centering
    \includegraphics[width=0.75\linewidth]{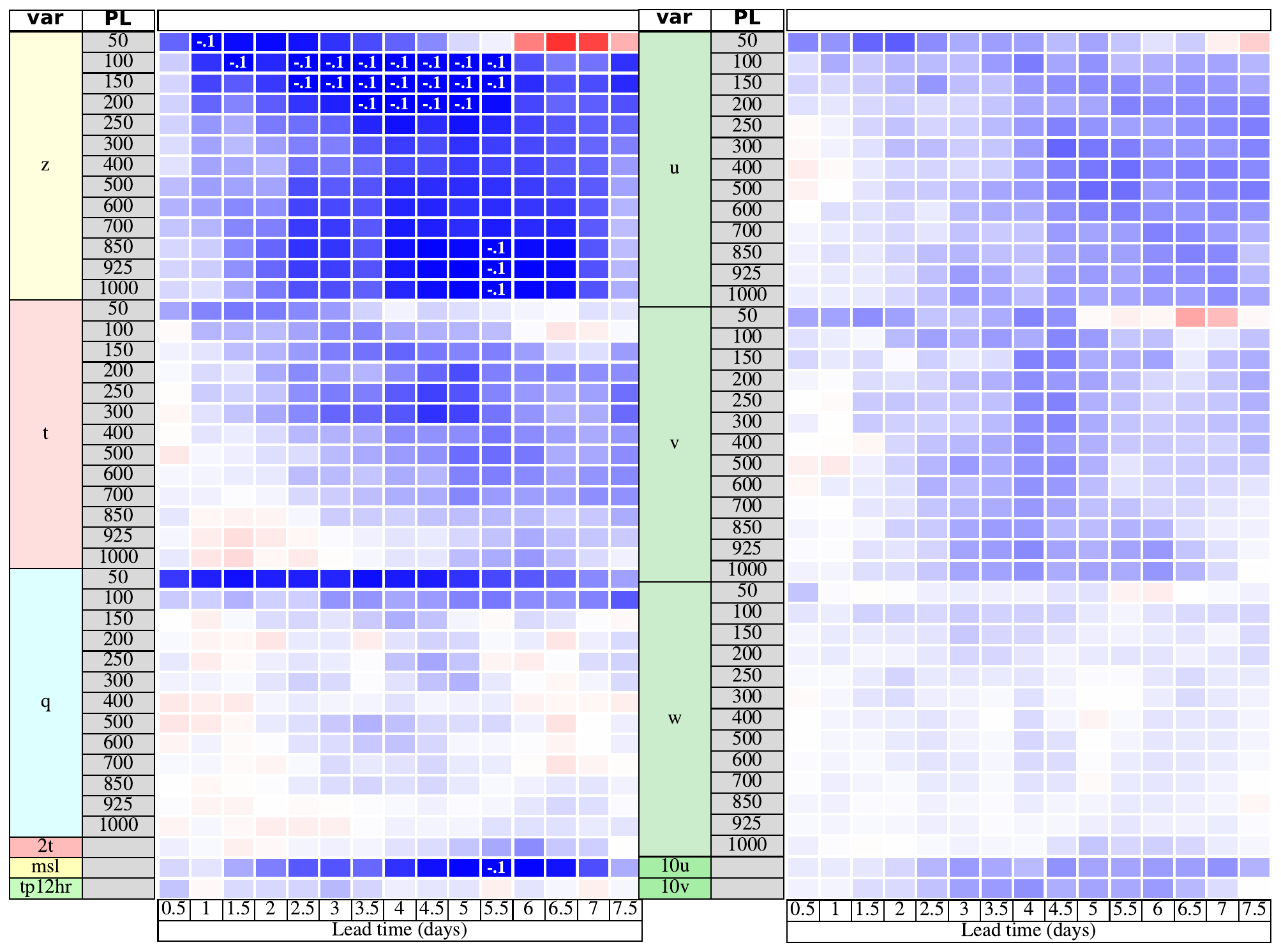}
    
    \caption{Per-variable CRPS relative improvements of the best RMMD-distilled model, relative to the GenCast teacher. Colorbars are scaled so to a maximum improvement/degradation of +5/-5\%.}
    \label{fig:crps_mmron}
\end{figure}

We analyse the performance of the best RMMD-distilled model separately for each weather variable. In Figure \ref{fig:crps_mmron}, we show the per-variable comparison of the GenCast distilled model compared to the teacher.
The distilled model with CRPS is better than the teacher on almost all weather variables and lead times, except humidity at small lead times (<2days)


\subsection{Making the strongest MMD baseline}

\begin{figure}[ht]
    \centering
    \includegraphics[width=0.8\linewidth]{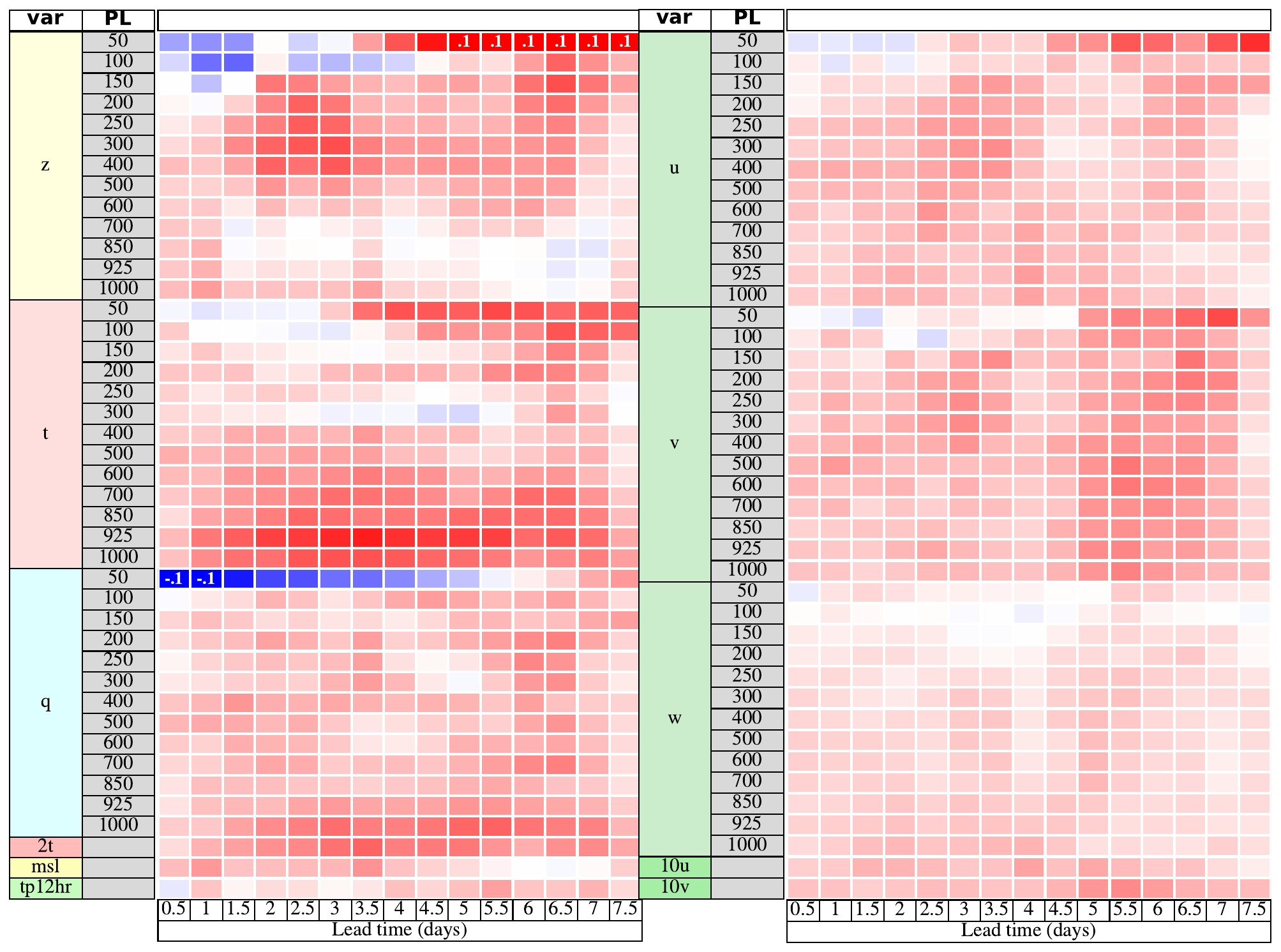}
    \caption{Per-variable CRPS improvement of an MMD-distilled model, without any change to the MMD parameters, compared to the GenCast teacher.}
    \label{fig:crps_mmd_init_vs_gencast}
\end{figure}

Without any change to MMD, the distilled model is worse than the teacher, as we can see in Figure \ref{fig:crps_mmd_init_vs_gencast}.

In DDPM, the posterior distribution $p_{cond}(x_s | x_t, \hat{x_0}^{(t)})$ is written:
\begin{align}
p_\text{cond}(x_s | x_t, \hat{x_0}^{(t)}) &= \mathcal{N}(\alpha_s \hat{x_0}^{(t)} + \sqrt{1-\alpha_s^2 - \gamma_{s, t}^2} \hat{\epsilon}, \gamma_{s,t}^2 I) \\
\text{with} \quad \hat{\epsilon} &= (x_t - \alpha_t \hat{x_0}^{(t)}) / \sigma_t \quad
\text{and} \quad \gamma_{s,t} = \eta \frac{\sigma_s}{\sigma_t} \sqrt{1 - \frac{\alpha_t^2}{\alpha_s^2}}
\end{align}

With $\eta=1$ corresponding to the classical DDPM posterior distribution and $\eta=0$ being deterministic sampling, which is not compatible with Moment Matching Distillation, since there is no distribution.

We found that using $\eta=0.5$ brings large improvements, as presented in Table \ref{tab:eta}. This parameter is used during sampling as well as during computing $x'_s$ from $x_t$ and $\hat{x}_0$ in the training loss of MMD. Increasing this parameter decreases the effect of stochasticity during the sampling loop. We found that can see that reducing this parameter reduces dispersion of the forecasts measured by spread-skill ratio.

However, decreasing this parameter further did not bring additional gains, which we hypothesize is because MMD relies on $x'_s$ being a non-deterministic function of $x_t$ and $\hat{x}_0$.

We also observed that changing the sampling schedule parameter $\rho$ from $\rho=7$ to $\rho=100$ (corresponding roughly to a uniform schedule in log-SNR parametrization) during distillation also improves CRPS scores and reduces under-dispersion.

\begin{table}[ht]
    \centering
    \begin{tabular}{l|r|r}
        & \makecell{CRPS \\ improv.(↑)} & \makecell{Win \\ rate(↑)} \\
        \hline
        Teacher & 0\% & N/A \\
        Plain MMD & -1.32\% & 4.9\% \\
        MMD w/ $\eta=0.5$ & 0.22\% & 50.7\% \\
        MMD (best) & 0.82\% & 75.0\% \\

    \end{tabular}
    \caption{Impact of parameter $\eta$ on CRPS improvement. Using $\eta=0.5$ improves performance a lot. In addition, using $\rho=100$ results in the best score for MMD.}
    \label{tab:eta}
\end{table}


\subsection{Churn analysis}

\begin{figure}[ht]
    \centering
    \includegraphics[width=1.0\linewidth]{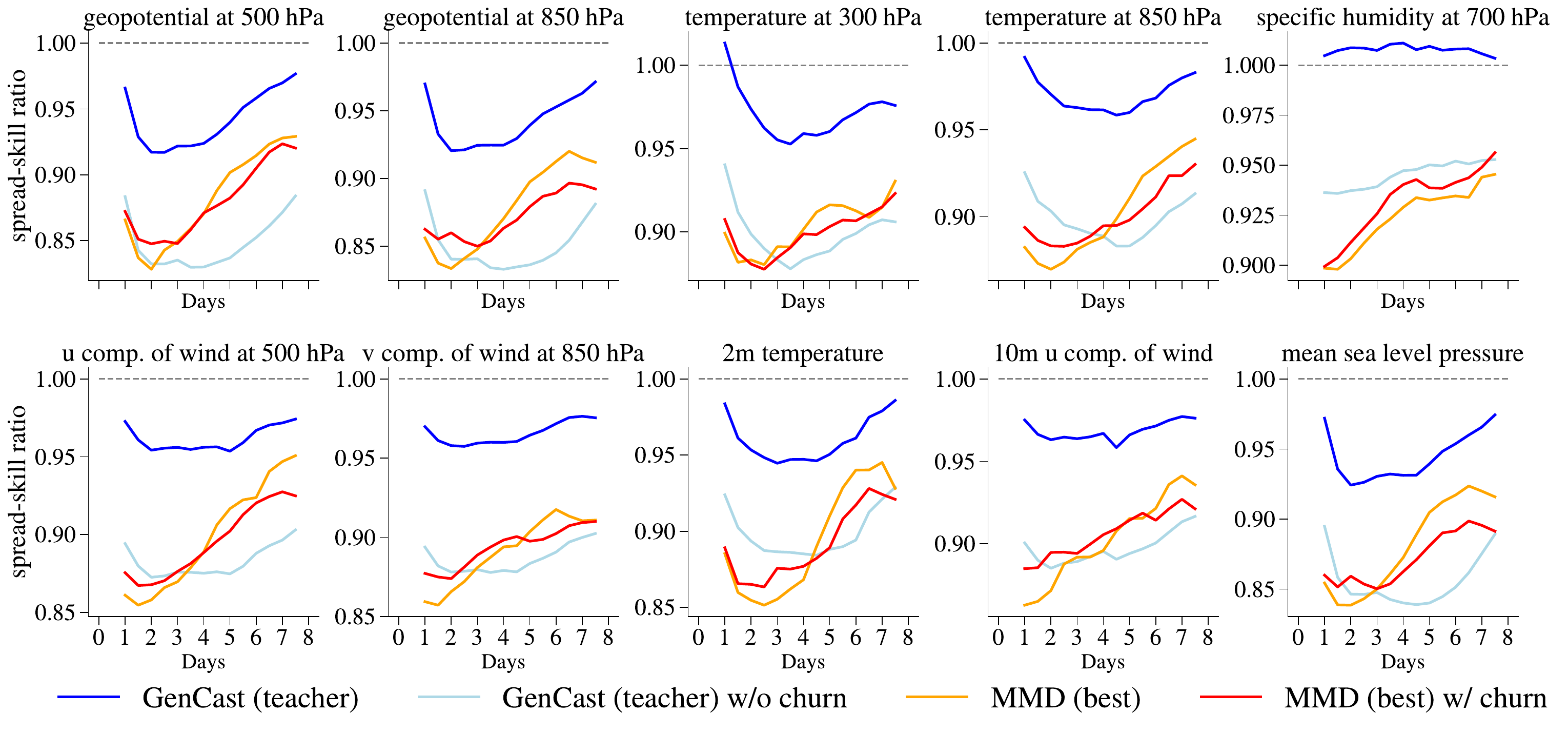}
    \caption{Spread-skill ratio of GenCast with and without inflated churn, and MMD-distilled model with and without inflated churn. We observe that the churn technique used in EDM is ineffective with MMD: It does not improve spread skill ratio and deteriorates CRPS scores.}
    \label{fig:ssr_churn}
\end{figure}

GenCast uses EDM sampling algorithm \citep{karras2022elucidating}, which is a deterministic sampling algorithm with stochastic churn on top. In theory, churn is not needed to sample from the correct distribution. In practice, GenCast uses $S_{noise} = 1.05$, which empirically helps improve the quality of samples. We also found that GenCast without stochastic churn provides under-dispersed forecasts (without enough diversity), and that stochastic churn helps to increase the spread of forecasts and therefore their calibration. The effect is quantified in Table \ref{tab:churn} and separated per lead time and physical variable in Figure \ref{fig:ssr_churn}. Although RMMD is based on DDPM sampling, we can implement a variant of stochastic churn with the same characteristics as for EDM sampling. We evaluate this variant for MMD only, which motivated us to not use stochastic churn for both MMD and RMMD.

\begin{table}[ht]
    \centering
    \begin{tabular}{l|r|r}
        Name & \makecell{CRPS \\ improv.(↑)} & \makecell{Win \\ rate(↑)} \\
        \hline
        GenCast Teacher & 0\% & N/A \\
        GenCast Teacher w/o churn & -0.57\% & 14.2\% \\
        MMD & 0.82\% & 75.0\% \\
        MMD w/ churn & 0.11\% & 50.2\% \\

    \end{tabular}
    \caption{Churn is helpful for the GenCast teacher but not for the distilled model.}
    \label{tab:churn}
\end{table}

We make the following observations:

\begin{itemize}
    \item First, if we compare MMD to the GenCast teacher without churn, we see that the MMD model is more under-dispersive at short lead times (0-3 days) and better calibrated at longer lead times. 
    \item Second, we see that adding churn to the MMD model does not help to improve dispersion. We can also see that it degrades CRPS, in Table \ref{tab:churn}.
    \item Our method does help to significantly improve spread-skill ratio over MMD (see Figure \ref{fig:spskr_mmr}), and we believe it is a better way to fix the under-dispersion issue of GenCast than stochastic churn.
\end{itemize}

\subsection{RMMD improvement over MMD}

In this section, we present the per-variable CRPS improvement of the RMMD-distilled models compared to the MMD checkpoint trained during the first stage. Interestingly, the on-policy version of RMMD improves upon the MMD initialization on every variable, which is not the case of the off-policy version.

\begin{figure}[htbp]
    \centering
    \includegraphics[width=0.75\linewidth]{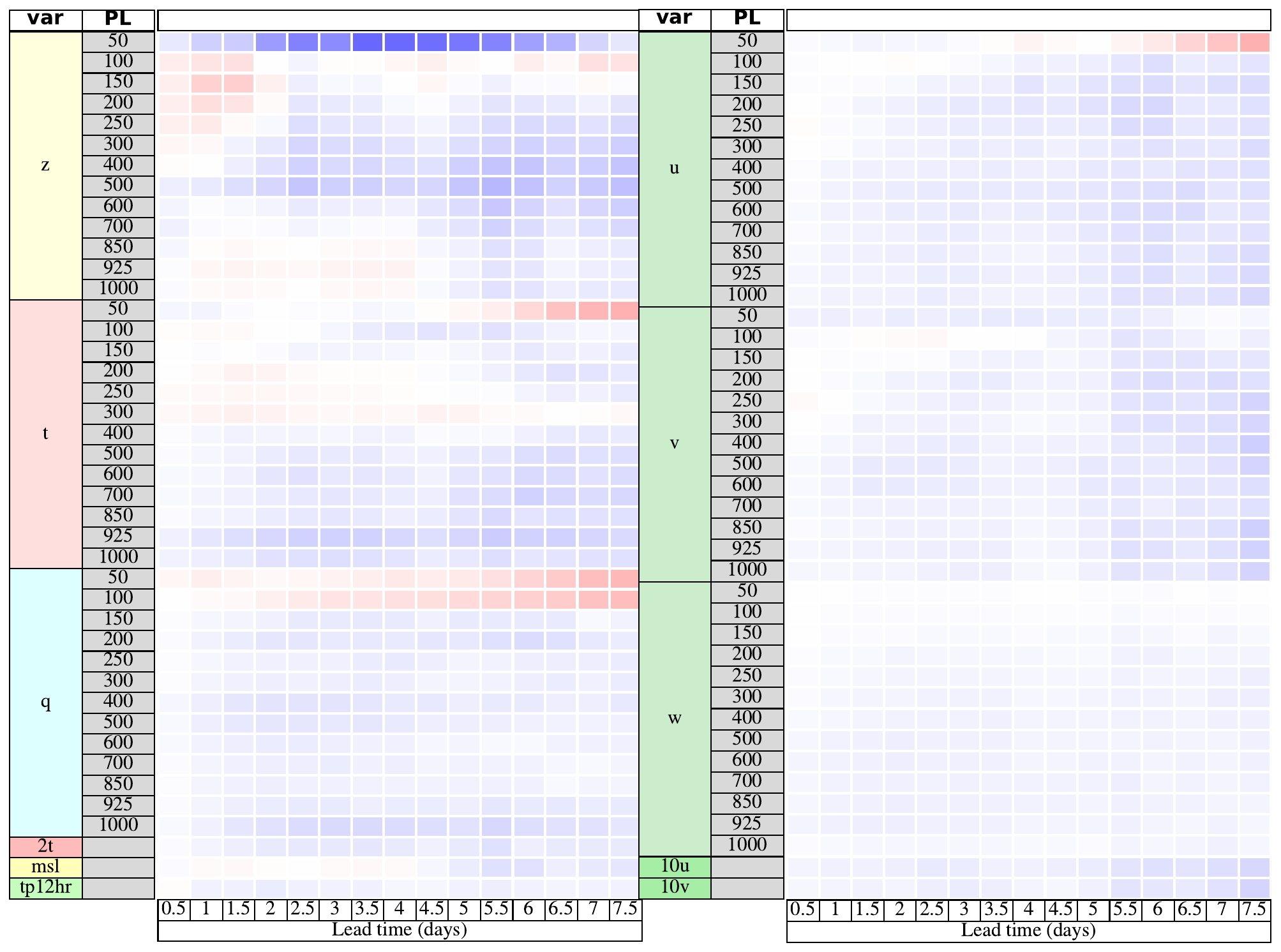}
    \includegraphics[width=0.75\linewidth]{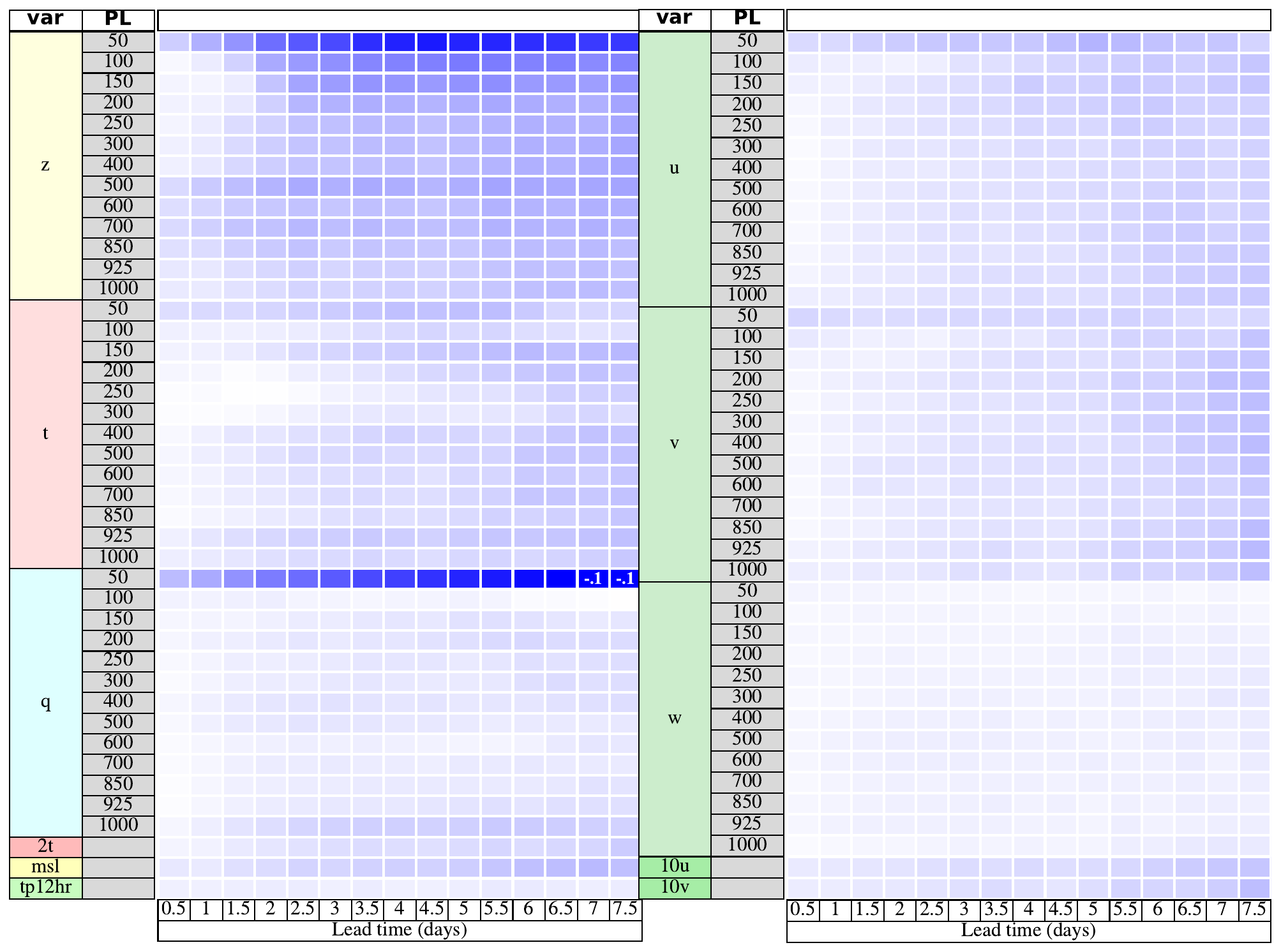}

    \caption{CRPS improvements of RMMD-distilled models compared to the MMD-distilled models in the first phase. Top: Off-policy version of RMMD. Bottom: RMMD (on-policy).}
    \label{fig:crps_vs_mmd}
\end{figure}

%% file: biblio.bib
@article{hoogeboom2023simple,
  title={Simple diffusion: End-to-end diffusion for high resolution images},
  author={Hoogeboom, Emiel and Heek, Jonathan and Salimans, Tim},
  journal={International Conference on Machine Learning},
  year={2023},
}

@article{hoogeboom2025simplerdiffusionsid215,
      title={Simpler Diffusion (SiD2): 1.5 FID on ImageNet512 with pixel-space diffusion}, 
      author={Emiel Hoogeboom and Thomas Mensink and Jonathan Heek and Kay Lamerigts and Ruiqi Gao and Tim Salimans},
      year={2025},
      journal={Conference on Computer Vision and Pattern Recognition}, 
}

@article{
song2021denoising,
title={Denoising Diffusion Implicit Models},
author={Jiaming Song and Chenlin Meng and Stefano Ermon},
journal={International Conference on Learning Representations},
year={2021},
}

@article{ho2020denoising,
  title={Denoising diffusion probabilistic models},
  author={Ho, Jonathan and Jain, Ajay and Abbeel, Pieter},
  journal={Advances in Neural Information Processing Systems},
  year={2020}
}

@article{de2025distributional,
  title={Distributional diffusion models with scoring rules},
  author={De Bortoli, Valentin and Galashov, Alexandre and Guntupalli, J Swaroop and Zhou, Guangyao and Murphy, Kevin and Gretton, Arthur and Doucet, Arnaud},
  journal={International Conference on Machine Learning},
  year={2025}
}

@inproceedings{esser2024scaling,
  title={Scaling rectified flow transformers for high-resolution image synthesis},
  author={Esser, Patrick and Kulal, Sumith and Blattmann, Andreas and Entezari, Rahim and M{\"u}ller, Jonas and Saini, Harry and Levi, Yam and Lorenz, Dominik and Sauer, Axel and Boesel, Frederic and others},
  booktitle={International Conference on Machine Learning},
  year={2024}
}

@article{song2023consistency,
  title={Consistency models},
  author={Song, Yang and Dhariwal, Prafulla and Chen, Mark and Sutskever, Ilya},
  journal={International Conference on Machine Learning},
  year={2023}
}

@article{luo2023diff,
  title={Diff-Instruct: A universal approach for transferring knowledge from pre-trained diffusion models},
  author={Luo, Weijian and Hu, Tianyang and Zhang, Shifeng and Sun, Jiacheng and Li, Zhenguo and Zhang, Zhihua},
  journal={Advances in Neural Information Processing Systems},
  volume={36},
  pages={76525--76546},
  year={2023}
}

@article{salimans2024multistep,
  title={Multistep distillation of diffusion models via moment matching},
  author={Salimans, Tim and Mensink, Thomas and Heek, Jonathan and Hoogeboom, Emiel},
  journal={Advances in Neural Information Processing Systems},
  year={2024}
}

@article{luo2024diff,
  title={Diff-Instruct++: Training one-step text-to-image generator model to align with human preferences},
  author={Luo, Weijian},
  journal={Transactions on Machine Learning Research},
  year={2024}
}

@article{
luo2025rewardinstruct,
title={Reward-Instruct: A Reward-Centric Approach to Fast Photo-Realistic Image Generation},
author={Yihong Luo and Tianyang Hu and Weijian Luo and Kenji Kawaguchi and Jing Tang},
journal={Advances in Neural Information Processing Systems},
year={2025},
}

@article{li2024reward,
  title={Reward guided latent consistency distillation},
  author={Li, Jiachen and Feng, Weixi and Chen, Wenhu and Wang, William Yang},
  journal={arXiv preprint arXiv:2403.11027},
  year={2024}
}

@article{ren2024hyper,
  title={Hyper-SD: Trajectory segmented consistency model for efficient image synthesis},
  author={Ren, Yuxi and Xia, Xin and Lu, Yanzuo and Zhang, Jiacheng and Wu, Jie and Xie, Pan and Wang, Xing and Xiao, Xuefeng},
  journal={Advances in Neural Information Processing Systems},
  year={2024}
}

@article{eyring2025noise,
  title={Noise Hypernetworks: Amortizing Test-Time Compute in Diffusion Models},
  author={Eyring, Luca and Karthik, Shyamgopal and Dosovitskiy, Alexey and Ruiz, Nataniel and Akata, Zeynep},
  journal={Advances in Neural Information Processing Systems},
  year={2025}
}

@article{marion2024implicit,
  title={Implicit diffusion: Efficient optimization through stochastic sampling},
  author={Marion, Pierre and Korba, Anna and Bartlett, Peter and Blondel, Mathieu and De Bortoli, Valentin and Doucet, Arnaud and Llinares-L{\'o}pez, Felipe and Paquette, Courtney and Berthet, Quentin},
  journal={Artificial Intelligence and Statistics},
  year={2025}
}

@article{clark2023directly,
  title={Directly fine-tuning diffusion models on differentiable rewards},
  author={Clark, Kevin and Vicol, Paul and Swersky, Kevin and Fleet, David J},
  journal={International Conference on Learning Representations},
  year={2024}
}

@article{xu2023imagereward,
  title={ImageReward: Learning and evaluating human preferences for text-to-image generation},
  author={Xu, Jiazheng and Liu, Xiao and Wu, Yuchen and Tong, Yuxuan and Li, Qinkai and Ding, Ming and Tang, Jie and Dong, Yuxiao},
  journal={Advances in Neural Information Processing Systems},
  year={2023}
}

@article{fan2023dpok,
  title={DPOK: Reinforcement learning for fine-tuning text-to-image diffusion models},
  author={Fan, Ying and Watkins, Olivia and Du, Yuqing and Liu, Hao and Ryu, Moonkyung and Boutilier, Craig and Abbeel, Pieter and Ghavamzadeh, Mohammad and Lee, Kangwook and Lee, Kimin},
  journal={Advances in Neural Information Processing Systems},
  year={2023}
}

@article{black2023training,
  title={Training diffusion models with reinforcement learning},
  author={Black, Kevin and Janner, Michael and Du, Yilun and Kostrikov, Ilya and Levine, Sergey},
  journal={International Conference on Learning Representations},
  year={2024}
}

@article{barratt2018note,
  title={A note on the inception score},
  author={Barratt, Shane and Sharma, Rishi},
  journal={arXiv preprint arXiv:1801.01973},
  year={2018}
}

@article{radford2021learning,
  title={Learning transferable visual models from natural language supervision},
  author={Radford, Alec and Kim, Jong Wook and Hallacy, Chris and Ramesh, Aditya and Goh, Gabriel and Agarwal, Sandhini and Sastry, Girish and Askell, Amanda and Mishkin, Pamela and Clark, Jack and others},
  journal={International Conference on Machine Learning},
  year={2021},
}

@article{song2020score,
  title={Score-based generative modeling through stochastic differential equations},
  author={Song, Yang and Sohl-Dickstein, Jascha and Kingma, Diederik P and Kumar, Abhishek and Ermon, Stefano and Poole, Ben},
  journal={International Conference on Learning Representations},
  year={2021}
}

@article{boffi2024flow,
  title={Flow map matching with stochastic interpolants: A mathematical framework for consistency models},
  author={Boffi, Nicholas M and Albergo, Michael S and Vanden-Eijnden, Eric},
  journal={arXiv preprint arXiv:2406.07507},
  year={2024}
}

@inproceedings{gengmeanflow2025,
  title={Mean Flows for One-step Generative Modeling},
  author={Geng, Zhengyang and Deng, Mingyang and Bai, Xingjian and Kolter, J Zico and He, Kaiming},
  booktitle={Advances in Neural Information Processing Systems},
  year={2025}
}

@article{uehara2024understanding,
  title={Understanding reinforcement learning-based fine-tuning of diffusion models: A tutorial and review},
  author={Uehara, Masatoshi and Zhao, Yulai and Biancalani, Tommaso and Levine, Sergey},
  journal={arXiv preprint arXiv:2407.13734},
  year={2024}
}

@inproceedings{deng2009imagenet,
  title={Imagenet: A large-scale hierarchical image database},
  author={Deng, Jia and Dong, Wei and Socher, Richard and Li, Li-Jia and Li, Kai and Fei-Fei, Li},
  booktitle={2009 IEEE Conference on Computer Vision and Pattern Recognition},
  year={2009},
  organization={Ieee}
}

@article{heusel2017gans,
  title={Gans trained by a two time-scale update rule converge to a local nash equilibrium},
  author={Heusel, Martin and Ramsauer, Hubert and Unterthiner, Thomas and Nessler, Bernhard and Hochreiter, Sepp},
  journal={Advances in Neural Information Processing Systems},
  volume={30},
  year={2017}
}

@inproceedings{song2024improved,
  title     = {Improved Techniques for Training Consistency Models},
  author    = {Yang Song and Prafulla Dhariwal},
  booktitle = {The Twelfth International Conference on Learning Representations},
  year      = {2024},
  url       = {https://openreview.net/forum?id=WNzy9bRDvG}
}

@inproceedings{kim2024consistency,
  title     = {Consistency Trajectory Models: Learning Probability Flow {ODE} Trajectory of Diffusion},
  author    = {Dongjun Kim and Chieh-Hsin Lai and Wei-Hsiang Liao and Naoki Murata and Yuhta Takida and Toshimitsu Uesaka and Yutong He and Yuki Mitsufuji and Stefano Ermon},
  booktitle = {The Twelfth International Conference on Learning Representations},
  year      = {2024},
  url       = {https://openreview.net/forum?id=ymjI8feDTD}
}

@inproceedings{frans2025one,
  title     = {One Step Diffusion via Shortcut Models},
  author    = {Kevin Frans and Danijar Hafner and Sergey Levine and Pieter Abbeel},
  booktitle = {The Thirteenth International Conference on Learning Representations},
  year      = {2025},
  url       = {https://openreview.net/forum?id=OlzB6LnXcS}
}

@misc{geng2025imf,
  title         = {Improved Mean Flows: On the Challenges of Fastforward Generative Models},
  author        = {Zhengyang Geng and Mingyang Deng and Xingjian Bai and J. Zico Kolter and Kaiming He},
  year          = {2025},
  eprint        = {2512.02012},
  archiveprefix = {arXiv},
  primaryclass  = {cs.CV},
  url           = {https://arxiv.org/abs/2512.02012}
}

@article{heek2024multistep,
  title={Multistep consistency models},
  author={Heek, Jonathan and Hoogeboom, Emiel and Salimans, Tim},
  journal={arXiv preprint arXiv:2403.06807},
  year={2024}
}

@article{geng2025mean,
  title={Mean flows for one-step generative modeling},
  author={Geng, Zhengyang and Deng, Mingyang and Bai, Xingjian and Kolter, J Zico and He, Kaiming},
  journal={arXiv preprint arXiv:2505.13447},
  year={2025}
}


%% file: gencast_bib.bib
@article{price2023gencast,
  title={GenCast: Diffusion-based ensemble forecasting for medium-range weather},
  author={Price, Ilan and Sanchez-Gonzalez, Alvaro and Alet, Ferran and Andersson, Tom R and El-Kadi, Andrew and Masters, Dominic and Ewalds, Timo and Stott, Jacklynn and Mohamed, Shakir and Battaglia, Peter and others},
  journal={arXiv preprint arXiv:2312.15796},
  year={2023}
}

@article{hersbach2020era5,
  title={The ERA5 global reanalysis},
  author={Hersbach, Hans and Bell, Bill and Berrisford, Paul and Hirahara, Shoji and Hor{\'a}nyi, Andr{\'a}s and Mu{\~n}oz-Sabater, Joaqu{\'\i}n and Nicolas, Julien and Peubey, Carole and Radu, Raluca and Schepers, Dinand and others},
  journal={Quarterly Journal of the Royal Meteorological Society},
  volume={146},
  number={730},
  pages={1999--2049},
  year={2020},
  publisher={Wiley Online Library}
}

@article{karras2022elucidating,
  title={Elucidating the design space of diffusion-based generative models},
  author={Karras, Tero and Aittala, Miika and Aila, Timo and Laine, Samuli},
  journal={Advances in Neural Information Processing Systems},
  year={2022}
}

@article{ferro2014fair,
  title={Fair scores for ensemble forecasts},
  author={Ferro, CAT},
  journal={Quarterly Journal of the Royal Meteorological Society},
  volume={140},
  number={683},
  pages={1917--1923},
  year={2014},
  publisher={Wiley Online Library}
}

@article{fortin2014should,
  title={Why should ensemble spread match the RMSE of the ensemble mean?},
  author={Fortin, Vincent and Abaza, Mabrouk and Anctil, Francois and Turcotte, Raphael},
  journal={Journal of Hydrometeorology},
  volume={15},
  number={4},
  pages={1708--1713},
  year={2014}
}

@misc{graphcast_github,
  author = {GoogleDeepMind},
  title = {Google DeepMind GraphCast and GenCast},
  year = {2025},
  publisher = {GitHub},
  journal = {GitHub repository},
  howpublished = {\url{https://github.com/google-deepmind/graphcast}},
}
